\documentclass[11pt]{article}

\usepackage[final]{acl}

\usepackage{times}
\usepackage{latexsym}

\usepackage[T1]{fontenc}

\usepackage[utf8]{inputenc}

\usepackage{microtype}

\usepackage{inconsolata}

\usepackage{graphicx}
\usepackage{amsmath}
\usepackage{amssymb}
\usepackage{subcaption}
\usepackage{longtable} %
\usepackage{stfloats}
\usepackage{cleveref}
\crefname{figure}{fig.}{figs.} %
\Crefname{figure}{Figure}{Figures} %

\usepackage{booktabs}
\usepackage{siunitx} %

\title{Navigating the Alignment-Calibration Trade-off: A Pareto-Superior Frontier via Model Merging
}

\author{
  Tiancheng Hu \qquad Benjamin Minixhofer \qquad Nigel Collier \\
  University of Cambridge \\
  \texttt{th656@cam.ac.uk}
}

\begin{document}
\maketitle

\begin{abstract}
The ``alignment tax'' of post-training is typically framed as a drop in task accuracy. We show it also involves a severe loss of calibration, making models overconfident, less reliable, and model outputs less diverse. We demonstrate that this trade-off can be navigated effectively via a simple post-hoc intervention: interpolating between a model's weights before and after alignment. Crucially, this is not a strict trade-off. We find that the process consistently reveals Pareto-optimal interpolations—models that improve accuracy beyond both parents while substantially recovering the calibration lost during alignment. Our work demonstrates that simple model merging provides a computationally efficient method for mitigating the full scope of the alignment tax, yielding models that are more capable and more reliable.\footnote{Our code is available \href{https://github.com/pitehu/model_merging}{here}.}

\end{abstract}

\section{Introduction}

Post-training is a double-edged sword. While it makes Large Language Models (LLMs) more helpful and safe, for instance by mitigating inherent social biases~\cite{Hu2025_socialidentitybias}, it also imposes a well-documented ``alignment tax'' by degrading performance on some benchmarks \cite{rlhf}. This tax, however, is often viewed narrowly through the lens of accuracy metrics. We argue this perspective overlooks a concurrent and equally critical problem: a severe degradation of model calibration. Alignment techniques are known to induce mode collapse, leading to overconfident, low-diversity outputs that signal a loss of the model's ability to represent uncertainty and produce diverse outputs~\cite{achiam2023gpt, kirk2024understanding,xiong2024can, wu2025generative}, even though being calibrated is key to user trust and actual usability of the model~\cite{xiong2024can,Steyvers2025}.

This paper connects these two phenomena. We propose a more holistic view of the alignment tax, framing it not just as a drop in accuracy, but as a broader degradation of model quality that encompasses both performance and calibration. By considering these issues in unison, we can devise more effective solutions. 

While recent work has sought to improve calibration by analyzing a model's intermediate representations~\cite{zhou2025beyond}, we show that post-hoc model merging is a simple, powerful, and computationally cheap remedy for this expanded alignment tax. By blending a well-calibrated pre-trained (PT) model with its aligned instruction-tuned (IT)\footnote{We use the term ``instruction tuning'' to refer to any alignment-related post-training, including but not limited to the narrowly defined instruction tuning} counterpart, we can navigate the trade-off between alignment and calibration. Our central discovery is that this is not necessarily a zero-sum game: we consistently identify ``sweet spot'' merges that \textbf{Pareto-dominate the instruction-tuned parent}, simultaneously achieving high accuracy while partially restoring calibration. Our work provides a practical path toward mitigating the full scope of the alignment tax, leading to models that are more capable, reliable, and diverse.

\section{Related Work}

\paragraph{Mitigating Alignment Tax}

A growing body of work seeks to mitigate the alignment tax~\cite{lu2024online,
fu-etal-2024-disperse, lin-etal-2024-mitigating,li2025preserving}, proposing novel regularization schemes or data curation techniques. Most closely related, \citet{lin-etal-2024-mitigating} use weight interpolation between pre-trained and RLHF-tuned models, but diagnose the tax exclusively through held-out perplexity and downstream accuracy; they do not study calibration, confidence distributions, or output diversity. Other merging works such as Rewarded Soups~\citep{rame2023rewarded} interpolate between models fine-tuned on \textit{different rewards}, rather than along the PT$\to$IT trajectory that defines the alignment tax. In contrast, our contribution is to show that calibration degradation is a central, previously uncharacterized component of the alignment tax, and that simple merging along the PT$\to$IT axis can recover it while maintaining or improving accuracy.

\paragraph{Model Merging}

Merging has been shown to be very effective at combining the capabilities of multiple specialized fine-tunes while largely retaining the constituent model’s strengths~\citep{utans1996weight,linear_merge,ilharco2023editing}. While the vast majority of merging research focuses on merging different fine-tunes~\citep{rame2023rewarded,
khalifa2024if}, some prior work aims to merge pre-trained and instruction-tuned models to enhance performance on specialized tasks~\citep{yu2024extend,wu2025shadow} or treat skills like instruction-following as transferable modules~\citep{cao2025paramdelta}; we discuss these strands of research in detail in Appendix~\ref{sec:appendix_rw}. In contrast, we merge PT and IT models with the goal of mitigating the alignment tax and restoring the model's fundamental calibration, unlike inference-time methods like Temperature Scaling, which adjust confidence but cannot improve accuracy.

\section{Experimental Setup}

We consider the Gemma-3 \citep{team2025gemma} and Qwen2.5 \citep{qwen2025qwen25technicalreport} model families, analyzing both their pre-trained (PT) base versions and officially released instruction-tuned (IT) counterparts. To explore the continuous space between a base model and its aligned version, we employ model merging to generate a series of interpolated models. We combine the weights of the PT model ($\theta_{\text{PT}}$) and IT model ($\theta_{\text{IT}}$) using a coefficient $\lambda \in [0, 1]$, where $\lambda=0$ recovers the pure PT model and $\lambda=1$ yields the pure IT model. Our primary results use Spherical Linear Interpolation (SLERP) \citep{slerp}, with linear interpolation and DARE-TIES \citep{linear_merge,dare} used to confirm robustness. All merging operations are performed using MergeKit \citep{mergekit}. Crucially, this is a post-hoc procedure that requires no additional training or gradient-based optimization and can be done without GPUs.

We evaluate models along two axes: \textbf{task performance}, measured by accuracy on a suite of challenging benchmarks (MMLU-Pro, GPQA, BBH, MATH, and IFEval), and \textbf{calibration}, measured by Expected Calibration Error (ECE; \citealp{Naeini2015obtaining, guo2017calibration}). This defines our \textit{alignment-calibration frontier}. All evaluations are conducted using the LM Evaluation Harness \citep{eval-harness}. We continue our discussion of related work in Appendix~\ref{sec:appendix_rw}.

\section{Results and Analysis}
\paragraph{The Calibration Cost of Instruction Tuning.}

While instruction tuning is a cornerstone of modern LLM development, it is not a panacea. We observe a consistent and significant trade-off between a model's capabilities and its calibration. Table~\ref{tab:calibration-cost-mmlu} quantifies this effect on the MMLU-Pro benchmark across a diverse set of models. The results reveal two patterns. First, accuracy on benchmarks like MMLU-Pro yields mixed results, with some models improving while others degrade—a known facet of the alignment tax. Second, calibration is universally degraded. ECE values consistently increase by an order of magnitude, signifying a severe rise in model overconfidence that undermines reliability. This ``calibration cost'' appears to be an inherent side effect of current instruction tuning methods. We see similar trends in the other datasets (Table~\ref{tab:comprehensive_results}).

\begin{table}[h!]
\centering
\resizebox{0.95\columnwidth}{!}{%
\begin{tabular}{l|cc|cc}
\toprule
\textbf{Model} & \multicolumn{2}{c|}{\textbf{Base}} & \multicolumn{2}{c}{\textbf{Instruct}} \\
\cmidrule(lr){2-3} \cmidrule(lr){4-5}
& Acc. (\%) & ECE $\downarrow$ & Acc. (\%) & ECE $\downarrow$ \\
\midrule
Gemma-3-1B & 11.2 & 0.07 & 14.2 & 0.66 \\
Gemma-3-4B & 27.9 & 0.02 & 29.8 & 0.64 \\
Gemma-3-12B & 42.4 & 0.02 & 39.8 & 0.53 \\
Gemma-3-27B & 49.4 & 0.04 & 47.8 & 0.48 \\
\midrule
Qwen2.5-1.5B & 28.7 & 0.06 & 28.1 & 0.33 \\
Qwen2.5-3B & 32.1 & 0.04 & 32.8 & 0.47 \\
Qwen2.5-7B & 43.6 & 0.06 & 43.1 & 0.45 \\
\bottomrule
\end{tabular}%
}
\caption{The Calibration Cost of Instruction Tuning on MMLU-Pro. Accuracy sees mixed results, while calibration is universally degraded.}
\label{tab:calibration-cost-mmlu}
\end{table}

\paragraph{Navigating the Frontier with Model Merging.}
Given this stark trade-off, we investigate model merging as a principled method to navigate the space between a base model's high calibration and an instruction-tuned model's alignment. Rather than treating the base and instruct models as discrete endpoints, merging allows us to trace the continuous frontier between them by varying a merge coefficient, $\lambda$. We found $\lambda > 1$ to catastrophically degrade performance~(Appendix~\ref{sec:appendix_amplifying}), so we constrain $\lambda \in [0, 1]$.

\begin{figure*}[t]
    \centering
    \includegraphics[width=0.999\linewidth]{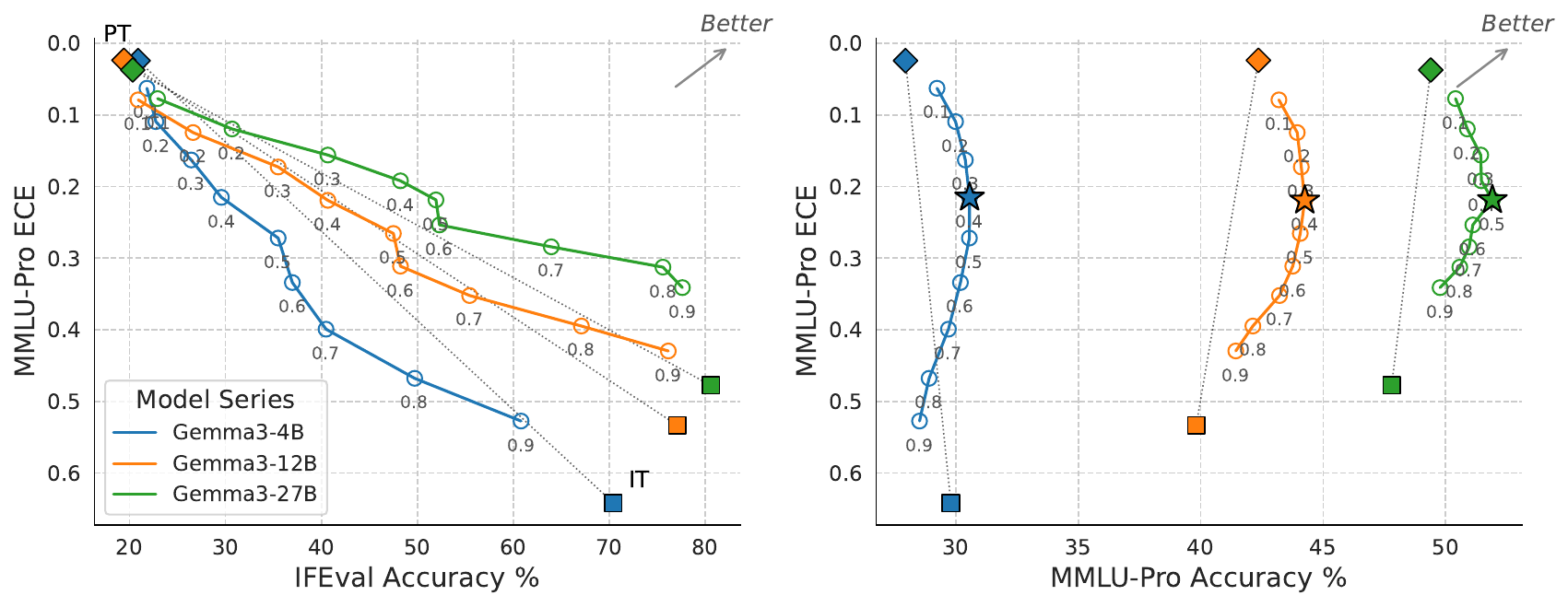}
\caption{The alignment-calibration frontier for Gemma-3 models. (\textbf{Left}) IFEval Accuracy vs. MMLU-Pro ECE. This cross-task view visualizes the fundamental trade-off between instruction-following and calibration on a knowledge task. (\textbf{Right}) MMLU-Pro Accuracy vs. ECE. Solid lines trace the performance of the PT and IT merges.}

\label{fig:gemma-frontier}

\label{fig:tradeoff-plot}
\end{figure*}

\begin{figure*}[tb!]
    \centering
    \includegraphics[width=0.48\textwidth]{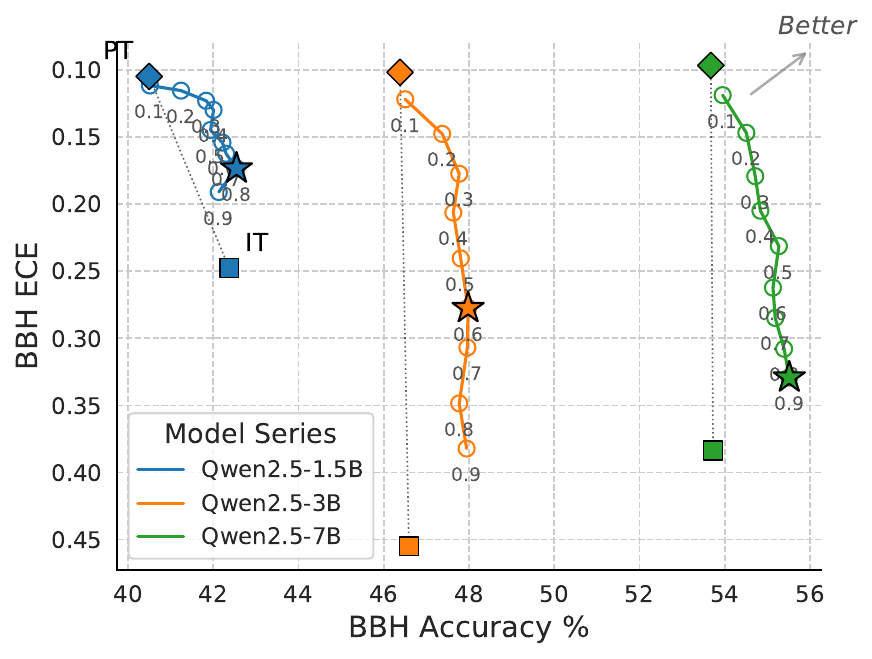}
    \hfill 
    \includegraphics[width=0.48\textwidth]{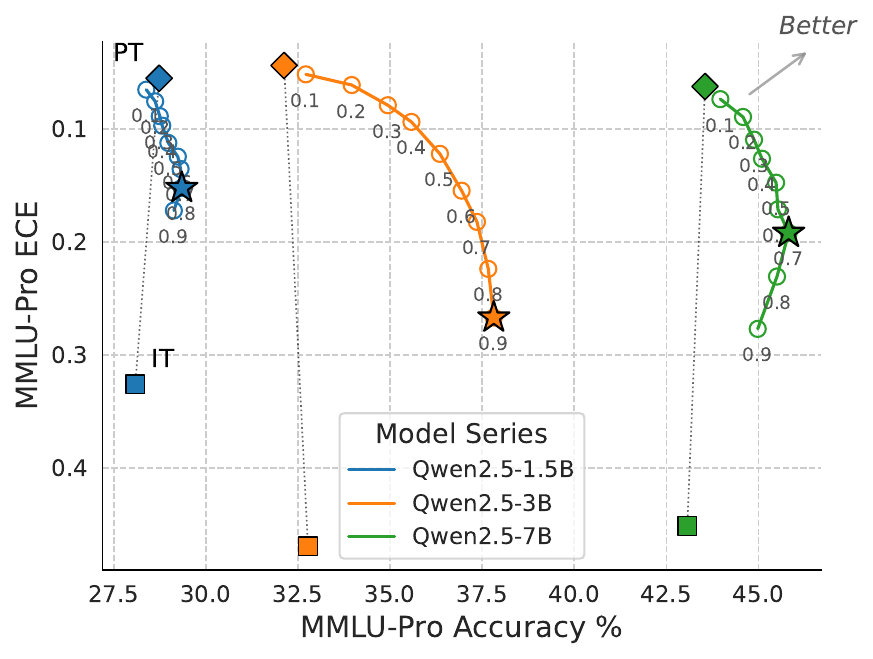}
        
    \caption{Performance vs. calibration for the Qwen2.5 model series on (\textbf{Left}) BBH and (\textbf{Right}) MMLU-Pro.}
    \label{fig:pareto-plots-combined}
\end{figure*}

Figure~\ref{fig:tradeoff-plot} illustrates this process for several models.
The left panel visualizes the ``cost of alignment.'' As the weight of the instruct model ($\lambda$) increases, alignment improves, but at a direct cost to calibration. This confirms that merging provides fine-grained control over this fundamental trade-off. The right panel reveals that model merging could create a Pareto-superior capability frontier: The path traced by the merged models (solid lines) strictly dominates the naïve linear interpolation between the base and instruct models (dotted lines). Crucially, this process uncovers an optimal merge coefficient ($\lambda^*$, marked by a star) that yields a model with accuracy comparable to or exceeding either of its parents, while simultaneously recovering a substantial portion of the calibration lost during instruction tuning.

\paragraph{General Trend Across Models, Datasets and Merge Methods.}
To demonstrate generality, we analyze the Qwen2.5 series on MMLU-Pro and BBH (Figure~\ref{fig:pareto-plots-combined}). The results are similar: merging creates a Pareto-superior frontier for both benchmarks. To further validate that these trends hold beyond the Gemma-3 and Qwen2.5 families, we conduct additional experiments on Llama-3.1-8B~\citep{grattafiori2024llama} using both SLERP and DARE-TIES. The same pattern emerges: at $\lambda=0.9$, the SLERP merge achieves 51.0\% on BBH (vs.\ PT 46.5\%, IT 50.7\%) with an ECE of only 0.074 (vs.\ IT 0.198), and on GPQA, the $\lambda=0.7$ merge reaches 33.0\% (exceeding both PT 31.5\% and IT 29.1\%) while keeping ECE at 0.065 (vs.\ IT 0.320). Full results across all three model families, sizes, and merging methods are reported in Table~\ref{tab:comprehensive_results}. This phenomenon is also robust across merging algorithms: Linear, SLERP, and DARE-TIES all trace distinct trajectories, yet consistently create frontiers dominating the PT-IT trade-off (Figure~\ref{fig:robustness_across_merging_method}).

\begin{figure*}[tb!]
    \centering
    \begin{subfigure}[b]{0.44\textwidth}
        \centering
        \includegraphics[width=\textwidth]{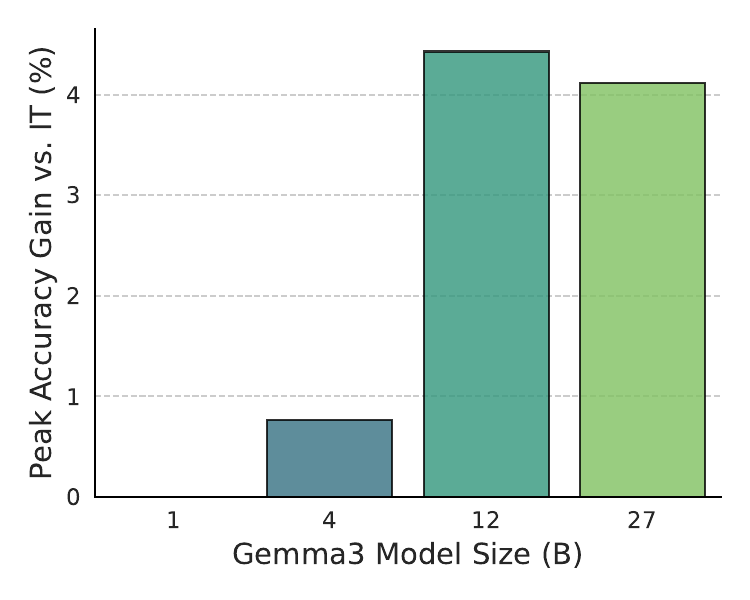}
        \caption{Peak accuracy gain vs.\ model size.}
        \label{fig:scaling_a}
    \end{subfigure}
    \hfill
    \begin{subfigure}[b]{0.44\textwidth}
        \centering
        \includegraphics[width=\textwidth]{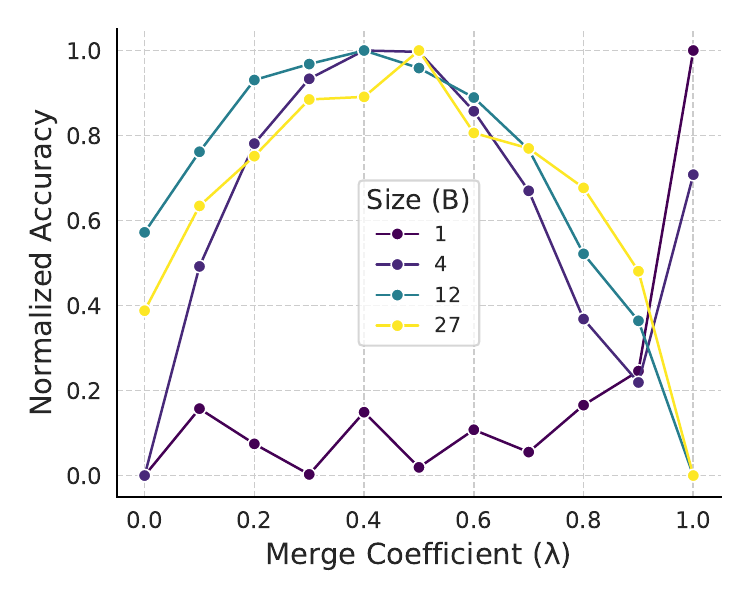}
        \caption{Normalized accuracy vs.\ merge coefficient ($\lambda$).}
        \label{fig:scaling_b}
    \end{subfigure}
    
    \vspace{0.5em}
    
    \begin{subfigure}[b]{0.44\textwidth}
        \centering
        \includegraphics[width=\textwidth]{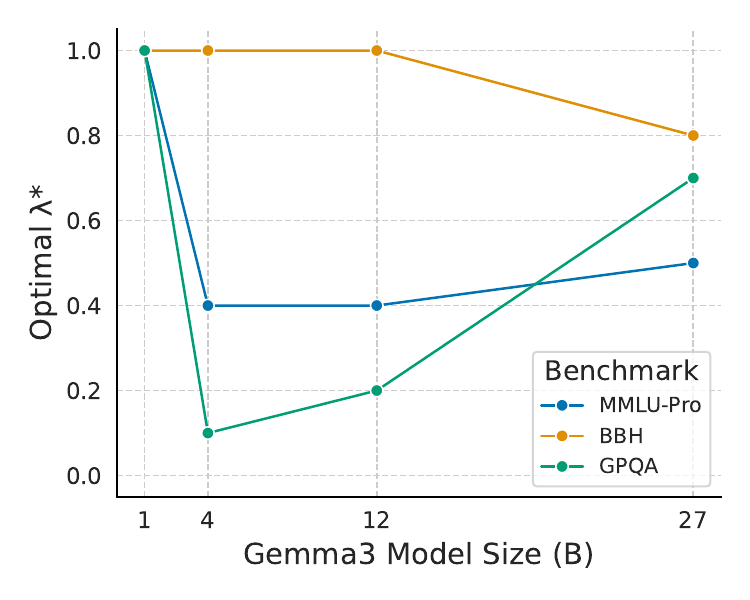}
        \caption{Optimal merge coefficient ($\lambda^*$) vs.\ model size.}
        \label{fig:scaling_c}
    \end{subfigure}
    \hfill
    \begin{subfigure}[b]{0.48\textwidth}
        \centering
        \includegraphics[width=\textwidth]{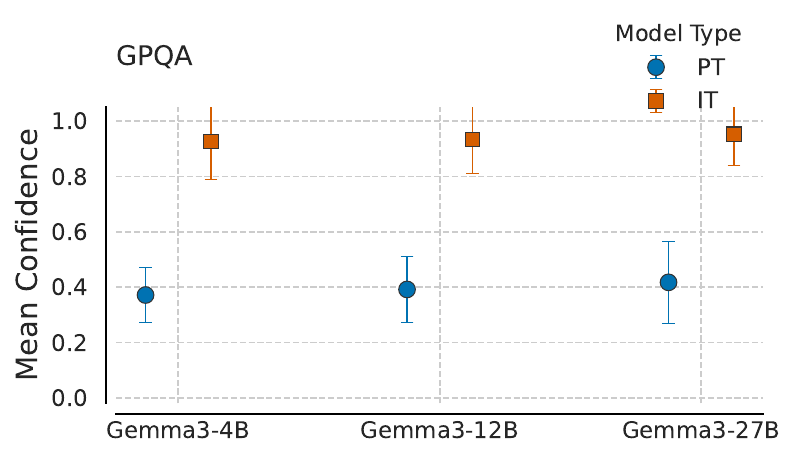}
        \caption{Mean prediction confidence of PT and IT models on GPQA.}
        \label{fig:confidence-plot}
    \end{subfigure}
    
    \caption{Scaling trends and mechanistic analysis of model merging for the Gemma3 family on MMLU-Pro. (a--c) The payoff, robustness, and predictability of merging all improve with model scale. (d) Instruction tuning inflates prediction confidence without a commensurate accuracy gain, driving the observed miscalibration.}
    \label{fig:scaling_and_mechanism}
    \label{fig:scaling_trends}
\end{figure*}

\paragraph{Merging Becomes More Effective and Robust at Scale.}
Beyond improving individual model performance, we find that the benefits and robustness of merging increase dramatically with model scale. Figure~\ref{fig:scaling_and_mechanism} illustrates this across the Gemma3 family on the MMLU-Pro benchmark. The peak accuracy gain from merging over the instruction-tuned parent grows substantially with model size (Figure \ref{fig:scaling_a}); while the gain is marginal for smaller models, it exceeds 4 percentage points for the 12B and 27B models. Furthermore, the process of finding an optimal merge becomes more robust. The performance landscape for larger models is a smooth, concave curve, indicating that performance is not highly sensitive to the exact merge coefficient (Figure \ref{fig:scaling_b}). In contrast, smaller models can exhibit more volatile behavior, making the choice of $\lambda$ more critical. Finally, the optimal merge strategy becomes more predictable at scale (Figure \ref{fig:scaling_c}). The optimal coefficient ($\lambda^*$) for tasks like MMLU-Pro and GPQA converges towards a stable value (around 0.4-0.5), and a simple default of $\lambda=0.5$ generalizes well across benchmarks.

\paragraph{Alignment Induces Calibration Error via Confidence Inflation.}

To understand the root cause of this calibration degradation, we investigate the underlying changes in model predictions. While accuracy on knowledge-intensive tasks often changes only marginally after instruction tuning (Table~\ref{tab:calibration-cost-mmlu}), we observe a dramatic shift in model confidence. As illustrated for the challenging GPQA benchmark in Figure~\ref{fig:confidence-plot}, the mean prediction confidence of instruction-tuned models surges from around 40\% to over 90\%. This sharp inflation of confidence, without a commensurate improvement in accuracy, is the direct driver of the observed catastrophic miscalibration. We provide a geometric interpretation of why merging reverses this effect through Linear Mode Connectivity in Appendix~\ref{sec:appendix_lmc}.

\paragraph{Restoring Calibration via Merging Improves Generative Diversity and Fidelity.}

\begin{figure}[h!]
    \centering
    \includegraphics[width=0.96\linewidth]{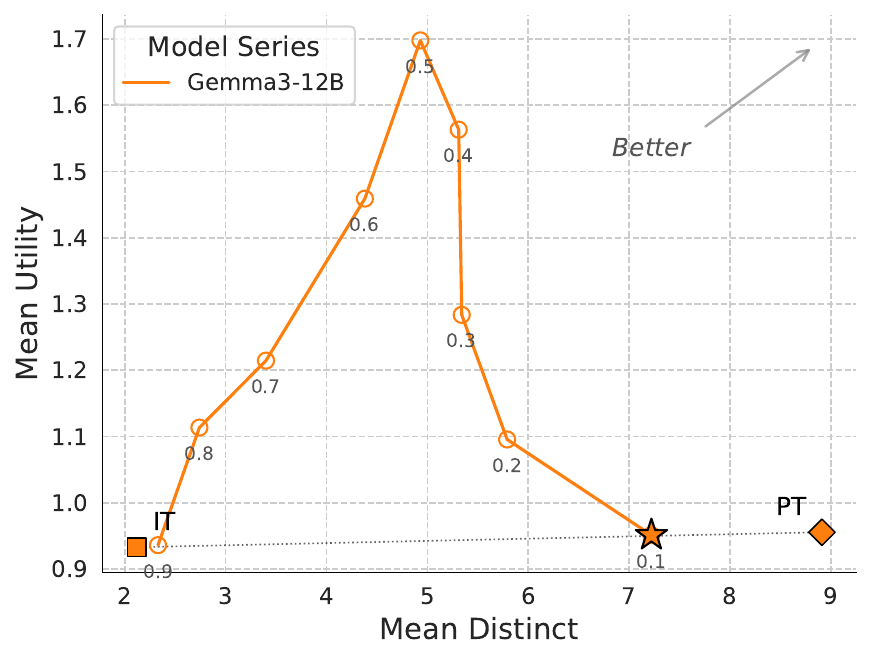}
    \caption{The trade-off between generation utility and distinctness on NoveltyBench (curated subset) for Gemma3-12B merges. The path traces merges from the PT to the IT parent.}
    \label{fig:novelty_bench}
\end{figure}

The benefits of our merging approach extend beyond discriminative tasks. On NoveltyBench (curated subset) for generative diversity (Figure \ref{fig:novelty_bench}), merging again reveals a Pareto-superior frontier. It resolves the trade-off between the over-confident IT model's low diversity (mode collapse) and the PT model's diffuse outputs, yielding a ``sweet spot'' model with both high utility and creative diversity. This trend holds on SimBench for distributional prediction, where the optimal merged model (score: 20.4) substantially outperforms both PT (7.7) and IT (18.2) parents. These findings demonstrate that restoring calibration unlocks tangible performance improvements across diverse applications.

\section{Conclusion}
We demonstrate that the ``alignment tax'' extends beyond accuracy degradation to a severe loss of model calibration. By framing this as a fundamental trade-off, we show that simple model merging is a computationally cheap and highly effective method for navigating the alignment-calibration frontier. Our central finding is that this is not a zero-sum game: we consistently identify merged ``sweet spot'' models that Pareto-dominate their parents, simultaneously improving task performance (up to >4 percentage points accuracy) while partially restoring calibration. We have further shown that these improvements scale favorably with model size, and are due to post-training causing overconfident predictions. This work provides a practical path toward developing LLMs that are both more capable and more reliable.

\section*{Limitations}

A key prerequisite for our approach is full access to the model weights. Consequently, our findings are most directly relevant to the open-source ecosystem, as the merging techniques we explore cannot be applied to proprietary, closed-source models that are accessible only through APIs.

Furthermore, our work focuses on demonstrating the alignment-calibration trade-off and its navigability using simple, computationally inexpensive merging techniques like linear interpolation and SLERP. We did not perform an exhaustive hyperparameter search for more complex methods (e.g., varying densities in DARE-TIES) nor did we explore more sophisticated merging strategies. For instance, hierarchical merging, where different merge coefficients are applied to different layers or modules, could offer finer-grained control over the trade-off and potentially unlock even better performance.

However, the primary goal of this work was to establish the existence of a Pareto-superior frontier using the most straightforward methods possible. The remarkable effectiveness of these simple approaches underscores the fundamental nature of our findings and highlights these more complex techniques as promising and important avenues for future research.

Finally, while we provide a geometric interpretation of our results through Linear Mode Connectivity (Appendix~\ref{sec:appendix_lmc}), we do not offer a formal mathematical theory for why Pareto-superior points, where merged models exceed both parents in accuracy, consistently emerge. Developing such a theory, potentially drawing on loss landscape geometry or connections to Bayesian model averaging, is an important direction for future work.

\section*{Ethical Considerations}

The primary goal of this research is to improve the reliability of LLMs by restoring the calibration lost during alignment, leading to more trustworthy systems. A well-calibrated model is less likely to be confidently wrong, which is a positive contribution to AI safety.

However, merging an instruction-tuned (IT) model with its base pre-trained (PT) counterpart inherently re-introduces weights from a model that has not undergone full safety fine-tuning. This process could potentially dilute or compromise the safety guardrails, such as refusal to generate harmful content, that were instilled during the alignment process. To quantify this risk, we conduct safety evaluations using ToxiGen and WMDP across the full $\lambda$ sweep (see Appendix~\ref{sec:appendix_safety}). Our results show that while toxicity increases as we move toward the base model, hazardous knowledge scores remain stable, and importantly, no merged model is less safe than the publicly available base model itself.

Our experiments are designed to illustrate the fundamental trade-off between alignment and calibration, not to prescribe a universally safe deployment strategy. We strongly caution practitioners that applying this technique requires careful evaluation. Any ``sweet spot'' model identified through merging must not only be assessed for accuracy and calibration but must also undergo rigorous and comprehensive safety testing to ensure it does not regress on critical safety benchmarks. The ultimate goal is to find a balance that enhances reliability without undermining the essential safety alignment of the model.

AI assistants were used for coding assistance and for copy-editing the paper.

\section*{Acknowledgements}
T.H is supported by Gates Cambridge Trust (grant OPP1144 from the Bill \& Melinda Gates Foundation). The authors acknowledge the use of resources provided by the Isambard-AI National AI Research Resource (AIRR). Isambard-AI is operated by the University of Bristol and is funded by the UK Government’s Department for Science, Innovation and Technology (DSIT) via UK Research and Innovation; and the Science and Technology Facilities Council [ST/AIRR/I-A-I/1023].

\bibliography{custom}

\begin{thebibliography}{48}
\providecommand{\natexlab}[1]{#1}

\bibitem[{Achiam et~al.(2023)Achiam, Adler, Agarwal, Ahmad, Akkaya, Aleman, Almeida, Altenschmidt, Altman, Anadkat et~al.}]{achiam2023gpt}
Josh Achiam, Steven Adler, Sandhini Agarwal, Lama Ahmad, Ilge Akkaya, Florencia~Leoni Aleman, Diogo Almeida, Janko Altenschmidt, Sam Altman, Shyamal Anadkat, and 1 others. 2023.
\newblock Gpt-4 technical report.
\newblock \emph{arXiv preprint arXiv:2303.08774}.

\bibitem[{Cao et~al.(2025)Cao, Wu, Prasad, Tian, and Liu}]{cao2025paramdelta}
Sheng Cao, Mingrui Wu, Karthik Prasad, Yuandong Tian, and Zechun Liu. 2025.
\newblock \href {https://openreview.net/forum?id=vqbd2OQnGp} {Param\${\textbackslash}delta\$ for direct mixing: Post-train large language model at zero cost}.
\newblock In \emph{The Thirteenth International Conference on Learning Representations}.

\bibitem[{Deep et~al.(2024)Deep, Bhardwaj, and Poria}]{deep2024dellamerging}
Pala~Tej Deep, Rishabh Bhardwaj, and Soujanya Poria. 2024.
\newblock \href {https://arxiv.org/abs/2406.11617} {Della-merging: Reducing interference in model merging through magnitude-based sampling}.
\newblock \emph{Preprint}, arXiv:2406.11617.

\bibitem[{Foret et~al.(2021)Foret, Kleiner, Mobahi, and Neyshabur}]{foret2021sharpnessaware}
Pierre Foret, Ariel Kleiner, Hossein Mobahi, and Behnam Neyshabur. 2021.
\newblock \href {https://openreview.net/forum?id=6Tm1mposlrM} {Sharpness-aware minimization for efficiently improving generalization}.
\newblock In \emph{International Conference on Learning Representations}.

\bibitem[{Fourrier et~al.(2024)Fourrier, Habib, Lozovskaya, Szafer, and Wolf}]{open-llm-leaderboard-v2}
Clémentine Fourrier, Nathan Habib, Alina Lozovskaya, Konrad Szafer, and Thomas Wolf. 2024.
\newblock Open llm leaderboard v2.
\newblock \url{https://huggingface.co/spaces/open-llm-leaderboard/open_llm_leaderboard}.

\bibitem[{Frankle et~al.(2020)Frankle, Dziugaite, Roy, and Carbin}]{DBLP:conf/icml/FrankleD0C20}
Jonathan Frankle, Gintare~Karolina Dziugaite, Daniel~M. Roy, and Michael Carbin. 2020.
\newblock \href {http://proceedings.mlr.press/v119/frankle20a.html} {Linear mode connectivity and the lottery ticket hypothesis}.
\newblock In \emph{Proceedings of the 37th International Conference on Machine Learning, {ICML} 2020, 13-18 July 2020, Virtual Event}, volume 119 of \emph{Proceedings of Machine Learning Research}, pages 3259--3269. {PMLR}.

\bibitem[{Fu et~al.(2024)Fu, Cai, Liu, Shi, and Yan}]{fu-etal-2024-disperse}
Tingchen Fu, Deng Cai, Lemao Liu, Shuming Shi, and Rui Yan. 2024.
\newblock \href {https://doi.org/10.18653/v1/2024.findings-acl.175} {Disperse-then-merge: Pushing the limits of instruction tuning via alignment tax reduction}.
\newblock In \emph{Findings of the Association for Computational Linguistics: ACL 2024}, pages 2967--2985, Bangkok, Thailand. Association for Computational Linguistics.

\bibitem[{Gao et~al.(2021)Gao, Tow, Biderman, Black, DiPofi, Foster, Golding, Hsu, McDonell, Muennighoff, Phang, Reynolds, Tang, Thite, Wang, Wang, and Zou}]{eval-harness}
Leo Gao, Jonathan Tow, Stella Biderman, Sid Black, Anthony DiPofi, Charles Foster, Laurence Golding, Jeffrey Hsu, Kyle McDonell, Niklas Muennighoff, Jason Phang, Laria Reynolds, Eric Tang, Anish Thite, Ben Wang, Kevin Wang, and Andy Zou. 2021.
\newblock \href {https://doi.org/10.5281/zenodo.5371628} {A framework for few-shot language model evaluation}.

\bibitem[{Goddard et~al.(2024)Goddard, Siriwardhana, Ehghaghi, Meyers, Karpukhin, Benedict, McQuade, and Solawetz}]{mergekit}
Charles Goddard, Shamane Siriwardhana, Malikeh Ehghaghi, Luke Meyers, Vladimir Karpukhin, Brian Benedict, Mark McQuade, and Jacob Solawetz. 2024.
\newblock \href {https://doi.org/10.18653/V1/2024.EMNLP-INDUSTRY.36} {Arcee's mergekit: {A} toolkit for merging large language models}.
\newblock In \emph{Proceedings of the 2024 Conference on Empirical Methods in Natural Language Processing: {EMNLP} 2024 - Industry Track, Miami, Florida, USA, November 12-16, 2024}, pages 477--485. Association for Computational Linguistics.

\bibitem[{Grattafiori et~al.(2024)Grattafiori, Dubey, Jauhri, Pandey, Kadian, Al-Dahle, Letman, Mathur, Schelten, Vaughan et~al.}]{grattafiori2024llama}
Aaron Grattafiori, Abhimanyu Dubey, Abhinav Jauhri, Abhinav Pandey, Abhishek Kadian, Ahmad Al-Dahle, Aiesha Letman, Akhil Mathur, Alan Schelten, Alex Vaughan, and 1 others. 2024.
\newblock The llama 3 herd of models.
\newblock \emph{arXiv preprint arXiv:2407.21783}.

\bibitem[{Guo et~al.(2017)Guo, Pleiss, Sun, and Weinberger}]{guo2017calibration}
Chuan Guo, Geoff Pleiss, Yu~Sun, and Kilian~Q Weinberger. 2017.
\newblock On calibration of modern neural networks.
\newblock In \emph{International conference on machine learning}, pages 1321--1330. PMLR.

\bibitem[{Hartvigsen et~al.(2022)Hartvigsen, Gabriel, Palangi, Sap, Ray, and Kamar}]{hartvigsen-etal-2022-toxigen}
Thomas Hartvigsen, Saadia Gabriel, Hamid Palangi, Maarten Sap, Dipankar Ray, and Ece Kamar. 2022.
\newblock \href {https://doi.org/10.18653/v1/2022.acl-long.234} {{T}oxi{G}en: A large-scale machine-generated dataset for adversarial and implicit hate speech detection}.
\newblock In \emph{Proceedings of the 60th Annual Meeting of the Association for Computational Linguistics (Volume 1: Long Papers)}, pages 3309--3326, Dublin, Ireland. Association for Computational Linguistics.

\bibitem[{Hendrycks et~al.(2021)Hendrycks, Burns, Kadavath, Arora, Basart, Tang, Song, and Steinhardt}]{hendrycks2021measuring}
Dan Hendrycks, Collin Burns, Saurav Kadavath, Akul Arora, Steven Basart, Eric Tang, Dawn Song, and Jacob Steinhardt. 2021.
\newblock \href {https://openreview.net/forum?id=7Bywt2mQsCe} {Measuring mathematical problem solving with the {MATH} dataset}.
\newblock In \emph{Thirty-fifth Conference on Neural Information Processing Systems Datasets and Benchmarks Track (Round 2)}.

\bibitem[{Hu et~al.(2026)Hu, Baumann, Lupo, Collier, Hovy, and R{\"o}ttger}]{hu2025simbench}
Tiancheng Hu, Joachim Baumann, Lorenzo Lupo, Nigel Collier, Dirk Hovy, and Paul R{\"o}ttger. 2026.
\newblock \href {https://openreview.net/forum?id=PL51SpN6ZJ} {Simbench: Benchmarking the ability of large language models to simulate human behaviors}.
\newblock In \emph{The Fourteenth International Conference on Learning Representations}.

\bibitem[{Hu et~al.(2025)Hu, Kyrychenko, Rathje, Collier, van~der Linden, and Roozenbeek}]{Hu2025_socialidentitybias}
Tiancheng Hu, Yara Kyrychenko, Steve Rathje, Nigel Collier, Sander van~der Linden, and Jon Roozenbeek. 2025.
\newblock \href {https://doi.org/10.1038/s43588-024-00741-1} {Generative language models exhibit social identity biases}.
\newblock \emph{Nature Computational Science}, 5(1):65--75.

\bibitem[{Ilharco et~al.(2023)Ilharco, Ribeiro, Wortsman, Schmidt, Hajishirzi, and Farhadi}]{ilharco2023editing}
Gabriel Ilharco, Marco~Tulio Ribeiro, Mitchell Wortsman, Ludwig Schmidt, Hannaneh Hajishirzi, and Ali Farhadi. 2023.
\newblock \href {https://openreview.net/forum?id=6t0Kwf8-jrj} {Editing models with task arithmetic}.
\newblock In \emph{The Eleventh International Conference on Learning Representations}.

\bibitem[{Izmailov et~al.(2018)Izmailov, Podoprikhin, Garipov, Vetrov, and Wilson}]{DBLP:conf/uai/IzmailovPGVW18}
Pavel Izmailov, Dmitrii Podoprikhin, Timur Garipov, Dmitry~P. Vetrov, and Andrew~Gordon Wilson. 2018.
\newblock \href {http://auai.org/uai2018/proceedings/papers/313.pdf} {Averaging weights leads to wider optima and better generalization}.
\newblock In \emph{Proceedings of the Thirty-Fourth Conference on Uncertainty in Artificial Intelligence, {UAI} 2018, Monterey, California, USA, August 6-10, 2018}, pages 876--885. {AUAI} Press.

\bibitem[{Khalifa et~al.(2024)Khalifa, Tan, Ahmadian, Hosking, Lee, Wang, {\"U}st{\"u}n, Sherborne, and Gall{\'e}}]{khalifa2024if}
Muhammad Khalifa, Yi-Chern Tan, Arash Ahmadian, Tom Hosking, Honglak Lee, Lu~Wang, Ahmet {\"U}st{\"u}n, Tom Sherborne, and Matthias Gall{\'e}. 2024.
\newblock If you can't use them, recycle them: Optimizing merging at scale mitigates performance tradeoffs.
\newblock \emph{arXiv preprint arXiv:2412.04144}.

\bibitem[{Kirk et~al.(2024)Kirk, Mediratta, Nalmpantis, Luketina, Hambro, Grefenstette, and Raileanu}]{kirk2024understanding}
Robert Kirk, Ishita Mediratta, Christoforos Nalmpantis, Jelena Luketina, Eric Hambro, Edward Grefenstette, and Roberta Raileanu. 2024.
\newblock \href {https://openreview.net/forum?id=PXD3FAVHJT} {Understanding the effects of {RLHF} on {LLM} generalisation and diversity}.
\newblock In \emph{The Twelfth International Conference on Learning Representations}.

\bibitem[{Li et~al.(2024)Li, Pan, Gopal, Yue, Berrios, Gatti, Li, Dombrowski, Goel, Mukobi, Helm{-}Burger, Lababidi, Justen, Liu, Chen, Barrass, Zhang, Zhu, Tamirisa, Bharathi, Herbert{-}Voss, Breuer, Zou, Mazeika, Wang, Oswal, Lin, Hunt, Tienken{-}Harder, Shih, Talley, Guan, Steneker, Campbell, Jokubaitis, Basart, Fitz, Kumaraguru, Karmakar, Tupakula, Varadharajan, Shoshitaishvili, Ba, Esvelt, Wang, and Hendrycks}]{DBLP:conf/icml/LiPGYBGLDGMHLJL24}
Nathaniel Li, Alexander Pan, Anjali Gopal, Summer Yue, Daniel Berrios, Alice Gatti, Justin~D. Li, Ann{-}Kathrin Dombrowski, Shashwat Goel, Gabriel Mukobi, Nathan Helm{-}Burger, Rassin Lababidi, Lennart Justen, Andrew~B. Liu, Michael Chen, Isabelle Barrass, Oliver Zhang, Xiaoyuan Zhu, Rishub Tamirisa, and 27 others. 2024.
\newblock \href {https://openreview.net/forum?id=xlr6AUDuJz} {The {WMDP} benchmark: Measuring and reducing malicious use with unlearning}.
\newblock In \emph{Forty-first International Conference on Machine Learning, {ICML} 2024, Vienna, Austria, July 21-27, 2024}. OpenReview.net.

\bibitem[{Li et~al.(2025)Li, Chen, Xu, Qin, Xiao, Luo, and Sun}]{li2025preserving}
Ziniu Li, Congliang Chen, Tian Xu, Zeyu Qin, Jiancong Xiao, Zhi-Quan Luo, and Ruoyu Sun. 2025.
\newblock \href {https://openreview.net/forum?id=NQEe7B7bSw} {Preserving diversity in supervised fine-tuning of large language models}.
\newblock In \emph{The Thirteenth International Conference on Learning Representations}.

\bibitem[{Lin et~al.(2024)Lin, Lin, Xiong, Diao, Liu, Zhang, Pan, Wang, Hu, Zhang, Dong, Pi, Zhao, Jiang, Ji, Yao, and Zhang}]{lin-etal-2024-mitigating}
Yong Lin, Hangyu Lin, Wei Xiong, Shizhe Diao, Jianmeng Liu, Jipeng Zhang, Rui Pan, Haoxiang Wang, Wenbin Hu, Hanning Zhang, Hanze Dong, Renjie Pi, Han Zhao, Nan Jiang, Heng Ji, Yuan Yao, and Tong Zhang. 2024.
\newblock \href {https://doi.org/10.18653/v1/2024.emnlp-main.35} {Mitigating the alignment tax of {RLHF}}.
\newblock In \emph{Proceedings of the 2024 Conference on Empirical Methods in Natural Language Processing}, pages 580--606, Miami, Florida, USA. Association for Computational Linguistics.

\bibitem[{Lu et~al.(2024)Lu, Yu, Huang, Fan, Lin, and Zhou}]{lu2024online}
Keming Lu, Bowen Yu, Fei Huang, Yang Fan, Runji Lin, and Chang Zhou. 2024.
\newblock Online merging optimizers for boosting rewards and mitigating tax in alignment.
\newblock \emph{arXiv preprint arXiv:2405.17931}.

\bibitem[{Ma et~al.(2024)Ma, Wang, Hu, Haensch, Hedderich, Plank, and Kreuter}]{ma-etal-2024-potential}
Bolei Ma, Xinpeng Wang, Tiancheng Hu, Anna-Carolina Haensch, Michael~A. Hedderich, Barbara Plank, and Frauke Kreuter. 2024.
\newblock \href {https://doi.org/10.18653/v1/2024.findings-emnlp.513} {The potential and challenges of evaluating attitudes, opinions, and values in large language models}.
\newblock In \emph{Findings of the Association for Computational Linguistics: EMNLP 2024}, pages 8783--8805, Miami, Florida, USA. Association for Computational Linguistics.

\bibitem[{Maddox et~al.(2019)Maddox, Izmailov, Garipov, Vetrov, and Wilson}]{DBLP:conf/nips/MaddoxIGVW19}
Wesley~J. Maddox, Pavel Izmailov, Timur Garipov, Dmitry~P. Vetrov, and Andrew~Gordon Wilson. 2019.
\newblock \href {https://proceedings.neurips.cc/paper/2019/hash/118921efba23fc329e6560b27861f0c2-Abstract.html} {A simple baseline for bayesian uncertainty in deep learning}.
\newblock In \emph{Advances in Neural Information Processing Systems 32: Annual Conference on Neural Information Processing Systems 2019, NeurIPS 2019, December 8-14, 2019, Vancouver, BC, Canada}, pages 13132--13143.

\bibitem[{Naeini et~al.(2015)Naeini, Cooper, and Hauskrecht}]{Naeini2015obtaining}
Mahdi~Pakdaman Naeini, Gregory~F. Cooper, and Milos Hauskrecht. 2015.
\newblock Obtaining well calibrated probabilities using bayesian binning.
\newblock In \emph{Proceedings of the Twenty-Ninth AAAI Conference on Artificial Intelligence}, AAAI'15, page 2901–2907. AAAI Press.

\bibitem[{Ouyang et~al.(2022)Ouyang, Wu, Jiang, Almeida, Wainwright, Mishkin, Zhang, Agarwal, Slama, Ray, Schulman, Hilton, Kelton, Miller, Simens, Askell, Welinder, Christiano, Leike, and Lowe}]{rlhf}
Long Ouyang, Jeff Wu, Xu~Jiang, Diogo Almeida, Carroll~L. Wainwright, Pamela Mishkin, Chong Zhang, Sandhini Agarwal, Katarina Slama, Alex Ray, John Schulman, Jacob Hilton, Fraser Kelton, Luke Miller, Maddie Simens, Amanda Askell, Peter Welinder, Paul Christiano, Jan Leike, and Ryan Lowe. 2022.
\newblock Training language models to follow instructions with human feedback.
\newblock In \emph{Proceedings of the 36th International Conference on Neural Information Processing Systems}, NIPS '22, Red Hook, NY, USA. Curran Associates Inc.

\bibitem[{Rame et~al.(2023)Rame, Couairon, Dancette, Gaya, Shukor, Soulier, and Cord}]{rame2023rewarded}
Alexandre Rame, Guillaume Couairon, Corentin Dancette, Jean-Baptiste Gaya, Mustafa Shukor, Laure Soulier, and Matthieu Cord. 2023.
\newblock \href {https://openreview.net/forum?id=lSbbC2VyCu} {Rewarded soups: towards pareto-optimal alignment by interpolating weights fine-tuned on diverse rewards}.
\newblock In \emph{Thirty-seventh Conference on Neural Information Processing Systems}.

\bibitem[{Rein et~al.(2024)Rein, Hou, Stickland, Petty, Pang, Dirani, Michael, and Bowman}]{rein2024gpqa}
David Rein, Betty~Li Hou, Asa~Cooper Stickland, Jackson Petty, Richard~Yuanzhe Pang, Julien Dirani, Julian Michael, and Samuel~R. Bowman. 2024.
\newblock \href {https://openreview.net/forum?id=Ti67584b98} {{GPQA}: A graduate-level google-proof q\&a benchmark}.
\newblock In \emph{First Conference on Language Modeling}.

\bibitem[{Shoemake(1985)}]{slerp}
Ken Shoemake. 1985.
\newblock \href {https://doi.org/10.1145/325334.325242} {Animating rotation with quaternion curves}.
\newblock In \emph{Proceedings of the 12th Annual Conference on Computer Graphics and Interactive Techniques}, SIGGRAPH '85, page 245–254, New York, NY, USA. Association for Computing Machinery.

\bibitem[{Steyvers et~al.(2025)Steyvers, Tejeda, Kumar, Belem, Karny, Hu, Mayer, and Smyth}]{Steyvers2025}
Mark Steyvers, Heliodoro Tejeda, Aakriti Kumar, Catarina Belem, Sheer Karny, Xinyue Hu, Lukas~W. Mayer, and Padhraic Smyth. 2025.
\newblock \href {https://doi.org/10.1038/s42256-024-00976-7} {What large language models know and what people think they know}.
\newblock \emph{Nature Machine Intelligence}, 7(2):221--231.

\bibitem[{Suzgun et~al.(2023)Suzgun, Scales, Sch{\"{a}}rli, Gehrmann, Tay, Chung, Chowdhery, Le, Chi, Zhou, and Wei}]{bbh}
Mirac Suzgun, Nathan Scales, Nathanael Sch{\"{a}}rli, Sebastian Gehrmann, Yi~Tay, Hyung~Won Chung, Aakanksha Chowdhery, Quoc~V. Le, Ed~H. Chi, Denny Zhou, and Jason Wei. 2023.
\newblock \href {https://doi.org/10.18653/V1/2023.FINDINGS-ACL.824} {Challenging big-bench tasks and whether chain-of-thought can solve them}.
\newblock In \emph{Findings of the Association for Computational Linguistics: {ACL} 2023, Toronto, Canada, July 9-14, 2023}, pages 13003--13051. Association for Computational Linguistics.

\bibitem[{Team et~al.(2025{\natexlab{a}})Team, Kamath, Ferret, Pathak, Vieillard, Merhej, Perrin, Matejovicova, Ram{\'e}, Rivi{\`e}re et~al.}]{team2025gemma}
Gemma Team, Aishwarya Kamath, Johan Ferret, Shreya Pathak, Nino Vieillard, Ramona Merhej, Sarah Perrin, Tatiana Matejovicova, Alexandre Ram{\'e}, Morgane Rivi{\`e}re, and 1 others. 2025{\natexlab{a}}.
\newblock Gemma 3 technical report.
\newblock \emph{arXiv preprint arXiv:2503.19786}.

\bibitem[{Team et~al.(2025{\natexlab{b}})Team, Yang, Yang, Zhang, Hui, Zheng, Yu, Li, Liu, Huang, Wei, Lin, Yang, Tu, Zhang, Yang, Yang, Zhou, Lin, Dang, Lu, Bao, Yang, Yu, Li, Xue, Zhang, Zhu, Men, Lin, Li, Tang, Xia, Ren, Ren, Fan, Su, Zhang, Wan, Liu, Cui, Zhang, and Qiu}]{qwen2025qwen25technicalreport}
Qwen Team, An~Yang, Baosong Yang, Beichen Zhang, Binyuan Hui, Bo~Zheng, Bowen Yu, Chengyuan Li, Dayiheng Liu, Fei Huang, Haoran Wei, Huan Lin, Jian Yang, Jianhong Tu, Jianwei Zhang, Jianxin Yang, Jiaxi Yang, Jingren Zhou, Junyang Lin, and 24 others. 2025{\natexlab{b}}.
\newblock \href {https://arxiv.org/abs/2412.15115} {Qwen2.5 technical report}.
\newblock \emph{Preprint}, arXiv:2412.15115.

\bibitem[{Utans(1996)}]{utans1996weight}
Joachim Utans. 1996.
\newblock Weight averaging for neural networks and local resampling schemes.
\newblock In \emph{Proceedings of the AAAI-96 Workshop on Integrating Multiple Learned Models}, pages 133--138.

\bibitem[{Wang et~al.(2023)Wang, Wei, Schuurmans, Le, Chi, Narang, Chowdhery, and Zhou}]{DBLP:conf/iclr/0002WSLCNCZ23}
Xuezhi Wang, Jason Wei, Dale Schuurmans, Quoc~V Le, Ed~H. Chi, Sharan Narang, Aakanksha Chowdhery, and Denny Zhou. 2023.
\newblock \href {https://openreview.net/forum?id=1PL1NIMMrw} {Self-consistency improves chain of thought reasoning in language models}.
\newblock In \emph{The Eleventh International Conference on Learning Representations}.

\bibitem[{Wang et~al.(2024)Wang, Ma, Zhang, Ni, Chandra, Guo, Ren, Arulraj, He, Jiang, Li, Ku, Wang, Zhuang, Fan, Yue, and Chen}]{wang2024mmlupro}
Yubo Wang, Xueguang Ma, Ge~Zhang, Yuansheng Ni, Abhranil Chandra, Shiguang Guo, Weiming Ren, Aaran Arulraj, Xuan He, Ziyan Jiang, Tianle Li, Max Ku, Kai Wang, Alex Zhuang, Rongqi Fan, Xiang Yue, and Wenhu Chen. 2024.
\newblock \href {https://openreview.net/forum?id=y10DM6R2r3} {{MMLU}-pro: A more robust and challenging multi-task language understanding benchmark}.
\newblock In \emph{The Thirty-eight Conference on Neural Information Processing Systems Datasets and Benchmarks Track}.

\bibitem[{Wortsman et~al.(2022)Wortsman, Ilharco, Gadre, Roelofs, Lopes, Morcos, Namkoong, Farhadi, Carmon, Kornblith, and Schmidt}]{linear_merge}
Mitchell Wortsman, Gabriel Ilharco, Samir~Yitzhak Gadre, Rebecca Roelofs, Raphael~Gontijo Lopes, Ari~S. Morcos, Hongseok Namkoong, Ali Farhadi, Yair Carmon, Simon Kornblith, and Ludwig Schmidt. 2022.
\newblock \href {https://proceedings.mlr.press/v162/wortsman22a.html} {Model soups: averaging weights of multiple fine-tuned models improves accuracy without increasing inference time}.
\newblock In \emph{International Conference on Machine Learning, {ICML} 2022, 17-23 July 2022, Baltimore, Maryland, {USA}}, volume 162 of \emph{Proceedings of Machine Learning Research}, pages 23965--23998. {PMLR}.

\bibitem[{Wu et~al.(2025{\natexlab{a}})Wu, Black, and Chandrasekaran}]{wu2025generative}
Fan Wu, Emily Black, and Varun Chandrasekaran. 2025{\natexlab{a}}.
\newblock \href {https://openreview.net/forum?id=yZ7sn9pyqb} {Generative monoculture in large language models}.
\newblock In \emph{The Thirteenth International Conference on Learning Representations}.

\bibitem[{Wu et~al.(2025{\natexlab{b}})Wu, Yang, Li, Hu, Wong, and Yang}]{wu2025shadow}
Taiqiang Wu, Runming Yang, Jiayi Li, Pengfei Hu, Ngai Wong, and Yujiu Yang. 2025{\natexlab{b}}.
\newblock Shadow-ft: Tuning instruct via base.
\newblock \emph{arXiv preprint arXiv:2505.12716}.

\bibitem[{Xiong et~al.(2024)Xiong, Hu, Lu, LI, Fu, He, and Hooi}]{xiong2024can}
Miao Xiong, Zhiyuan Hu, Xinyang Lu, YIFEI LI, Jie Fu, Junxian He, and Bryan Hooi. 2024.
\newblock \href {https://openreview.net/forum?id=gjeQKFxFpZ} {Can {LLM}s express their uncertainty? an empirical evaluation of confidence elicitation in {LLM}s}.
\newblock In \emph{The Twelfth International Conference on Learning Representations}.

\bibitem[{Yadav et~al.(2023)Yadav, Tam, Choshen, Raffel, and Bansal}]{yadav2023tiesmerging}
Prateek Yadav, Derek Tam, Leshem Choshen, Colin Raffel, and Mohit Bansal. 2023.
\newblock \href {https://openreview.net/forum?id=xtaX3WyCj1} {{TIES}-merging: Resolving interference when merging models}.
\newblock In \emph{Thirty-seventh Conference on Neural Information Processing Systems}.

\bibitem[{Yang et~al.(2026)Yang, Shen, Guo, Wang, Cao, Zhang, and Tao}]{yang2024model}
Enneng Yang, Li~Shen, Guibing Guo, Xingwei Wang, Xiaochun Cao, Jie Zhang, and Dacheng Tao. 2026.
\newblock \href {https://doi.org/10.1145/3787849} {Model merging in llms, mllms, and beyond: Methods, theories, applications, and opportunities}.
\newblock \emph{ACM Comput. Surv.}, 58(8).

\bibitem[{Yu et~al.(2024{\natexlab{a}})Yu, Yu, Yu, Huang, and Li}]{yu2024extend}
Le~Yu, Bowen Yu, Haiyang Yu, Fei Huang, and Yongbin Li. 2024{\natexlab{a}}.
\newblock Extend model merging from fine-tuned to pre-trained large language models via weight disentanglement.
\newblock \emph{arXiv preprint arXiv:2408.03092}.

\bibitem[{Yu et~al.(2024{\natexlab{b}})Yu, Yu, Yu, Huang, and Li}]{dare}
Le~Yu, Bowen Yu, Haiyang Yu, Fei Huang, and Yongbin Li. 2024{\natexlab{b}}.
\newblock Language models are super mario: absorbing abilities from homologous models as a free lunch.
\newblock In \emph{Proceedings of the 41st International Conference on Machine Learning}, ICML'24. JMLR.org.

\bibitem[{Zhang et~al.(2025)Zhang, Diddee, Holm, Liu, Liu, Samuel, Wang, and Ippolito}]{zhang2025noveltybench}
Yiming Zhang, Harshita Diddee, Susan Holm, Hanchen Liu, Xinyue Liu, Vinay Samuel, Barry Wang, and Daphne Ippolito. 2025.
\newblock \href {https://openreview.net/forum?id=XZm1ekzERf} {Noveltybench: Evaluating creativity and diversity in language models}.
\newblock In \emph{Second Conference on Language Modeling}.

\bibitem[{Zhou et~al.(2025)Zhou, Zhang, Hu, Li, Collier, Vuli{\'c}, and Korhonen}]{zhou2025beyond}
Ej~Zhou, Caiqi Zhang, Tiancheng Hu, Chengzu Li, Nigel Collier, Ivan Vuli{\'c}, and Anna Korhonen. 2025.
\newblock Beyond the final layer: Intermediate representations for better multilingual calibration in large language models.
\newblock \emph{arXiv preprint arXiv:2510.03136}.

\bibitem[{Zhou et~al.(2023)Zhou, Lu, Mishra, Brahma, Basu, Luan, Zhou, and Hou}]{zhou2023ifeval}
Jeffrey Zhou, Tianjian Lu, Swaroop Mishra, Siddhartha Brahma, Sujoy Basu, Yi~Luan, Denny Zhou, and Le~Hou. 2023.
\newblock \href {https://arxiv.org/abs/2311.07911} {Instruction-following evaluation for large language models}.
\newblock \emph{Preprint}, arXiv:2311.07911.

\end{thebibliography}
\clearpage
\appendix
\section{Task Description and Implementation Details}
\label{app:tasks}
We run all model inference in BF16. We adopt the standard configurations of the Open LLM Leaderboard \citep{open-llm-leaderboard-v2} for few-shot counts and prompting, but do not perform accuracy normalization, as raw accuracy is required for calibration calculations.

\paragraph{MMLU-Pro}~\cite{wang2024mmlupro}
An enhanced version of MMLU, MMLU-Pro increases difficulty by incorporating more reasoning-focused questions and expanding the choice set from four to ten, significantly reducing the chance of correct guesses. It spans 14 diverse domains, offering a more robust and discriminative evaluation of language understanding. We measure accuracy using a 5-shot setup.

\paragraph{GPQA}~\cite{rein2024gpqa}
GPQA is a graduate-level, ``Google-proof'' question-answering benchmark with questions authored by domain experts in biology, physics, and chemistry. It is designed to be extremely difficult for non-experts to solve, even with web access, providing a rigorous test of advanced reasoning. We measure accuracy using a 0-shot setup.
\paragraph{Big-Bench Hard (BBH)}~\cite{bbh}
A curated subset of 23 challenging tasks from the BIG-Bench suite where prior models performed below the average human-rater baseline. BBH focuses on tasks requiring complex, multi-step reasoning, such as logical deduction and multi-step arithmetic. We measure accuracy using a 3-shot setup.
\paragraph{MATH}~\cite{hendrycks2021measuring}
The MATH dataset evaluates mathematical problem-solving with problems from high school competitions. These problems require sophisticated reasoning beyond simple calculation. For our experiments, we focus exclusively on the most challenging problems, designated as Level 5. We measure exact match accuracy in a 4-shot setting.
\paragraph{IFEval}~\cite{zhou2023ifeval}
IFEval (Instruction-Following Evaluation) assesses an LLM's ability 
to adhere to explicit, verifiable instructions within a prompt. 
It automatically checks compliance with constraints on format 
(e.g., ``end your response with...''), length (e.g., ``write 
at least 400 words''), and content (e.g., ``mention `AI' 3 times''). 
We measure strict accuracy in a 0-shot setting.
\paragraph{NoveltyBench}~\cite{zhang2025noveltybench}
NoveltyBench evaluates a model's ability to generate diverse and novel outputs, counteracting the ``mode collapse'' phenomenon where models produce repetitive answers. It uses open-ended prompts and measures performance with metrics for distinctness (number of unique ideas) and utility (a combined score of novelty and quality).
\paragraph{SimBench}~\cite{hu2025simbench}
SimBench is a large-scale, standardized benchmark for evaluating how well LLMs simulate human behavior. It unifies 20 diverse social and behavioral science datasets to test a model's ability to predict group-level response distributions across various human populations and tasks, from moral dilemmas to economic choices. Measuring such complex, human-centric capabilities is a significant challenge~\cite{ma-etal-2024-potential}, and SimBench provides a concrete framework for doing so.

\section{Full Merging Results}

This section provides the comprehensive empirical data that underpins the analyses and conclusions presented in the main body of the paper. Table \ref{tab:comprehensive_results} offers a detailed breakdown of performance and calibration metrics across our entire suite of experiments, demonstrating the generality and robustness of our findings.

The table is organized by model family (Gemma-3, Qwen2.5 and Llama-3.1), model scale, and merging algorithm (SLERP, Linear, and DARE-TIES). For each configuration, we report results for the base Pre-Trained (PT) and Instruction-Tuned (IT) models, which serve as the endpoints ($\lambda=0$ and $\lambda=1$, respectively), alongside the series of merged models with the interpolation coefficient $\lambda$ varying from 0.1 to 0.9.

\section{Full Results for Scaling Analysis}
\label{sec:appendix_scaling}

\begin{figure*}[htbp]
    \centering
    \begin{subfigure}[b]{0.32\textwidth}
        \centering
        \includegraphics[width=\textwidth]{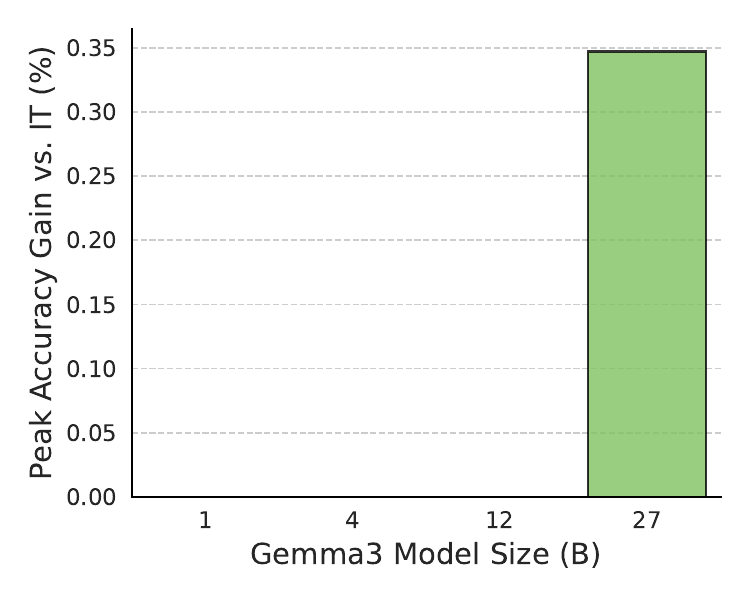}
    \end{subfigure}
    \hfill
    \begin{subfigure}[b]{0.32\textwidth}
        \centering
        \includegraphics[width=\textwidth]{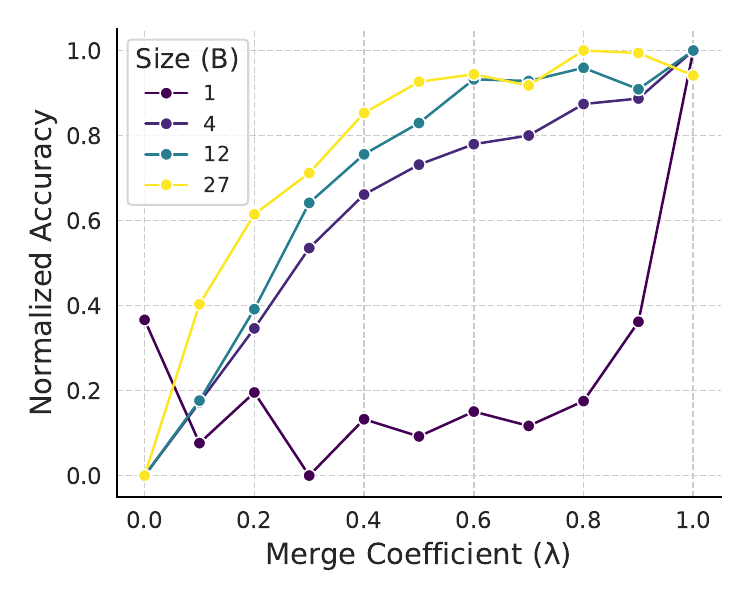}
    \end{subfigure}
    \hfill
    \begin{subfigure}[b]{0.32\textwidth}
        \centering
        \includegraphics[width=\textwidth]{figures/scaling_analysis/Gemma3/Gemma3_panel_c_lambda_convergence.pdf}
    \end{subfigure}
    \caption{Scaling trends for the Gemma3 family on the \textbf{BBH} benchmark.}
    \label{fig:appendix_gemma_bbh}
\end{figure*}

\begin{figure*}[htbp]
    \centering
    \begin{subfigure}[b]{0.32\textwidth}
        \centering
        \includegraphics[width=\textwidth]{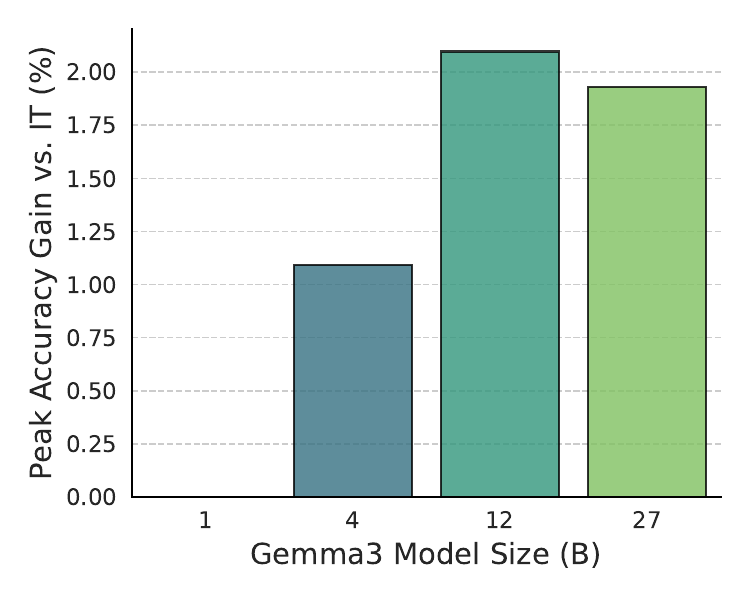}
    \end{subfigure}
    \hfill
    \begin{subfigure}[b]{0.32\textwidth}
        \centering
        \includegraphics[width=\textwidth]{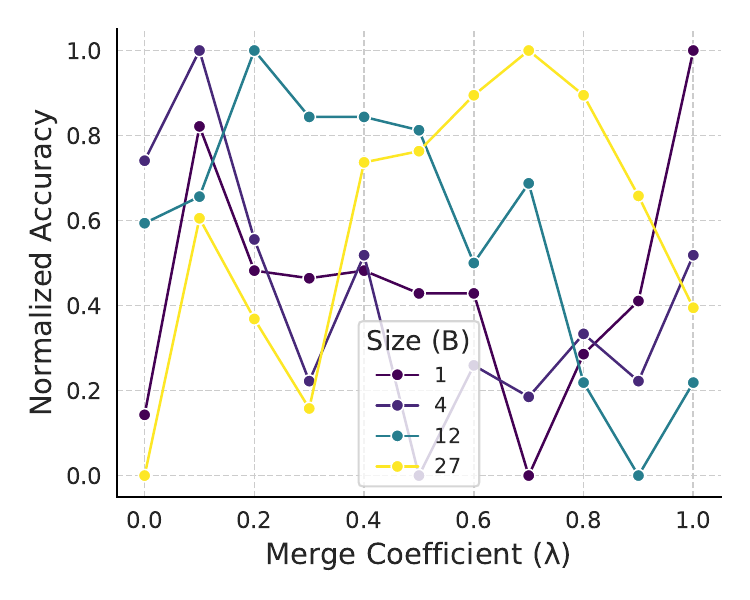}
    \end{subfigure}
    \hfill
    \begin{subfigure}[b]{0.32\textwidth}
        \centering
        \includegraphics[width=\textwidth]{figures/scaling_analysis/Gemma3/Gemma3_panel_c_lambda_convergence.pdf}
    \end{subfigure}
    \caption{Scaling trends for the Gemma3 family on the \textbf{GPQA} benchmark.}
    \label{fig:appendix_gemma_gpqa}
\end{figure*}

\begin{figure*}[htbp]
    \centering
    \begin{subfigure}[b]{0.32\textwidth}
        \centering
        \includegraphics[width=\textwidth]{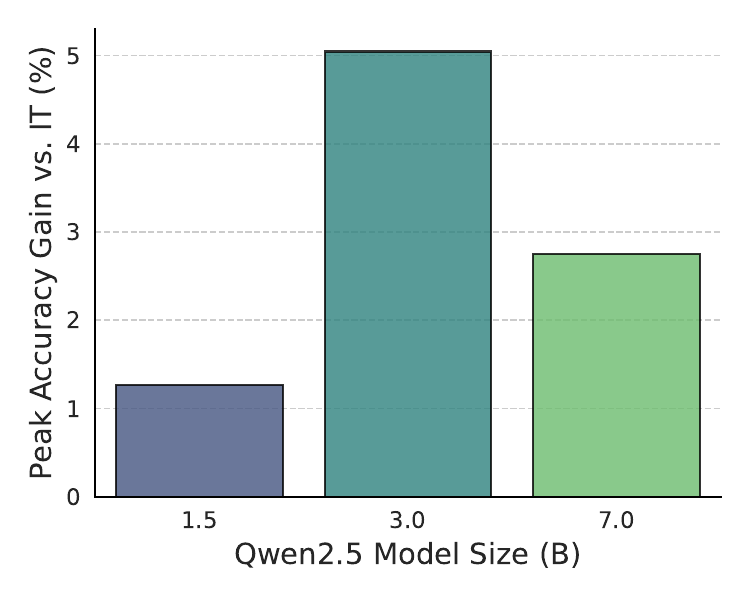}
    \end{subfigure}
    \hfill
    \begin{subfigure}[b]{0.32\textwidth}
        \centering
        \includegraphics[width=\textwidth]{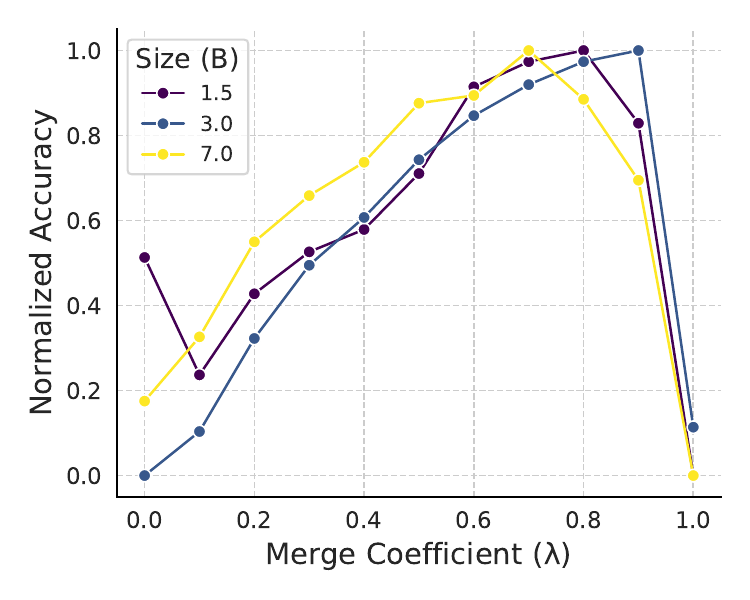}
    \end{subfigure}
    \hfill
    \begin{subfigure}[b]{0.32\textwidth}
        \centering
        \includegraphics[width=\textwidth]{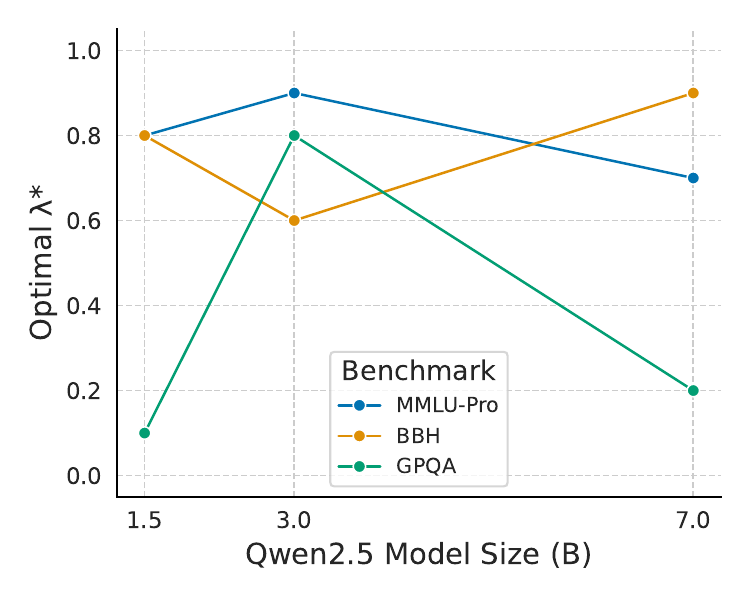}
    \end{subfigure}
    \caption{Scaling trends for the Qwen2.5 family on the \textbf{MMLU-Pro} benchmark.}
    \label{fig:appendix_qwen_mmlu}
\end{figure*}

\begin{figure*}[htbp]
    \centering
    \begin{subfigure}[b]{0.32\textwidth}
        \centering
        \includegraphics[width=\textwidth]{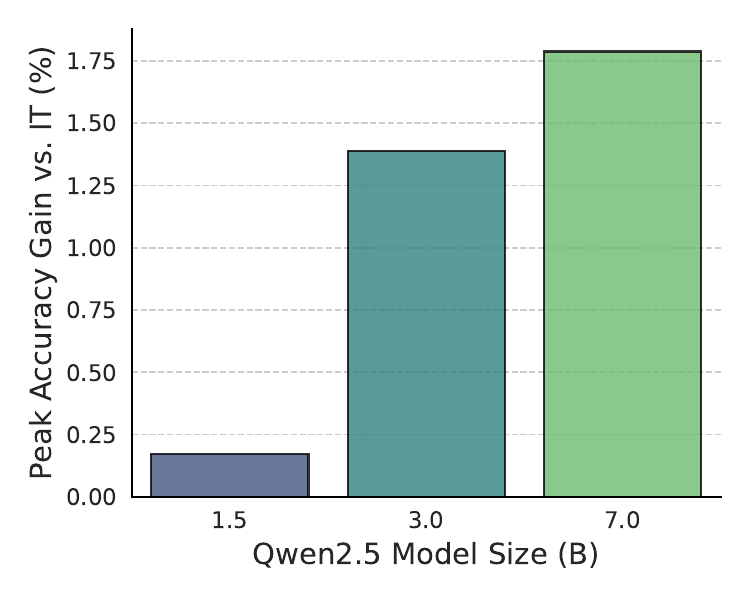}
    \end{subfigure}
    \hfill
    \begin{subfigure}[b]{0.32\textwidth}
        \centering
        \includegraphics[width=\textwidth]{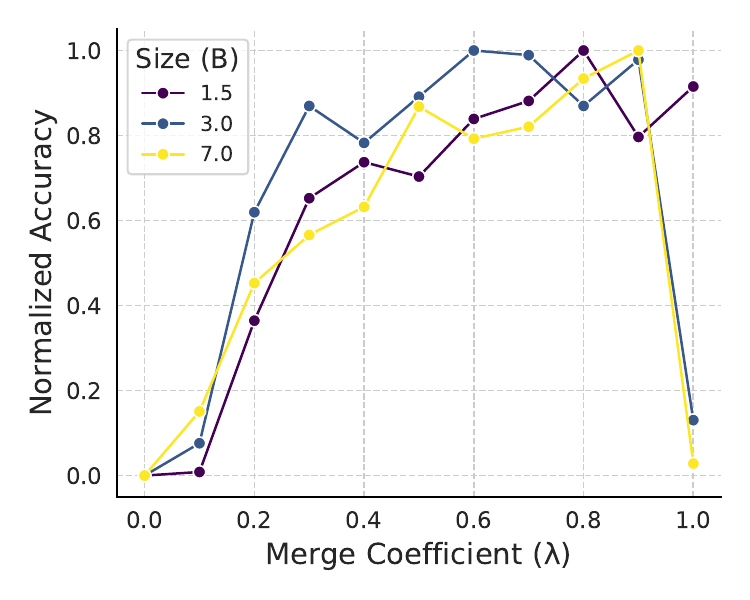}
    \end{subfigure}
    \hfill
    \begin{subfigure}[b]{0.32\textwidth}
        \centering
        \includegraphics[width=\textwidth]{figures/scaling_analysis/Qwen2.5/Qwen2.5_panel_c_lambda_convergence.pdf}
    \end{subfigure}
    \caption{Scaling trends for the Qwen2.5 family on the \textbf{BBH} benchmark.}
    \label{fig:appendix_qwen_bbh}
\end{figure*}

\begin{figure*}[htbp]
    \centering
    \begin{subfigure}[b]{0.32\textwidth}
        \centering
        \includegraphics[width=\textwidth]{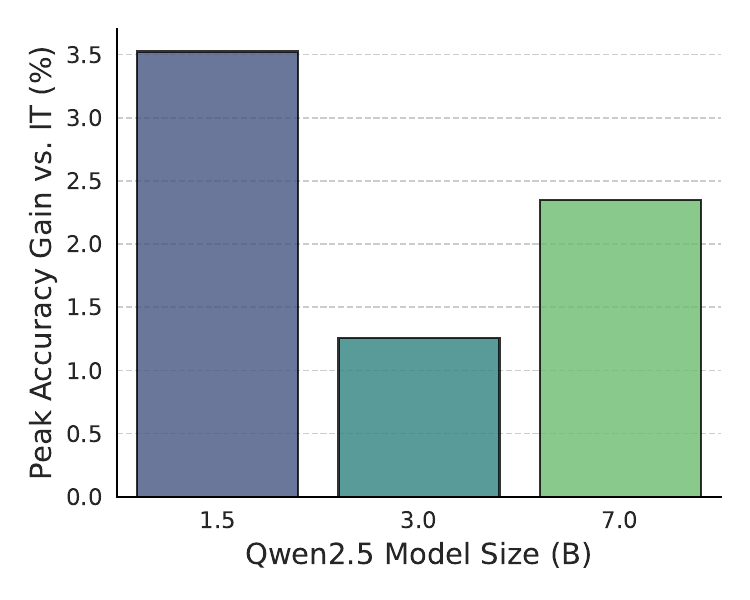}
        \caption{Payoff vs. Scale}
    \end{subfigure}
    \hfill
    \begin{subfigure}[b]{0.32\textwidth}
        \centering
        \includegraphics[width=\textwidth]{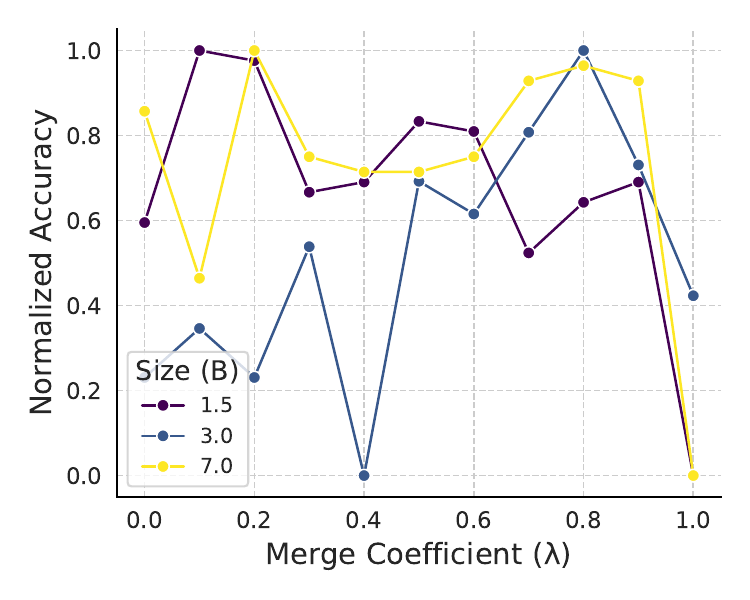}
        \caption{Robustness vs. Scale}
    \end{subfigure}
    \hfill
    \begin{subfigure}[b]{0.32\textwidth}
        \centering
        \includegraphics[width=\textwidth]{figures/scaling_analysis/Qwen2.5/Qwen2.5_panel_c_lambda_convergence.pdf}
        \caption{Predictability vs. Scale}
    \end{subfigure}
    \caption{Scaling trends for the Qwen2.5 family on the \textbf{GPQA} benchmark.}
    \label{fig:appendix_qwen_gpqa}
\end{figure*}

This appendix (\Cref{fig:appendix_gemma_bbh,fig:appendix_gemma_gpqa,fig:appendix_qwen_mmlu,fig:appendix_qwen_bbh,fig:appendix_qwen_gpqa}) provides the full set of figures supporting our scaling analysis in the main text. We demonstrate that the core trends observed for the Gemma3 family on MMLU-Pro - namely that the peak accuracy gain, performance landscape smoothness, and optimal $\lambda^*$ convergence of merging all increase with scale - are broadly consistent across the Qwen2.5 model family and other challenging benchmarks (BBH and GPQA). While the magnitude of the effects varies by task and model family, the overarching conclusion remains robust: model merging is an increasingly effective, stable, and practical technique for larger models.

\section{Additional Related Work and Background}
\label{sec:appendix_rw}
Model merging is the process of combining model parameters  $\theta_1, .., \theta_n$ into a single set of parameters $\theta_{\text{merge}}$, where $\theta_1, .., \theta_n$ are typically fine-tunes of the same base model. 
There exists a wide variety of model merging methods, ranging from simple (spherical) linear interpolation between the constituent model parameters~\citep{ilharco2023editing,mergekit} to sophisticated methods aiming to minimize interference between the merge constituents~\citep{dare,yadav2023tiesmerging,deep2024dellamerging}.\footnote{See \citet{yang2024model} for a detailed overview.} 

While the vast majority of merging research focuses on merging different fine-tuned models, some prior research investigates merging pre-trained and instruction-tuned models.

\paragraph{Shadow-FT~\citep{wu2025shadow}} proposes that additional fine-tuning of an instruction-tuned model (e.g., to specialize to a particular domain) can be improved by conducting the additional training on the base model, then merging this additionally trained model and the original instruction-tuned model.

\paragraph{Param $\Delta$~\citep{cao2025paramdelta}} suggests that an instruction-tuned model can be transferred to an updated base model (i.e., a newer base model version) by adding the instruction-tuning task vector $\theta_{\text{IT}} - \theta_{\text{PT}}$ to the new backbone weights $\theta_{\text{PT}'}$ analogous to Task Arithmetic~\citep{ilharco2023editing}.\footnote{Interestingly, this suggests that our method could also be used to interpolate between an instruction-tuned model and a \textit{newer version} of its base model. However, we do not investigate this further here.}

\paragraph{WIDEN~\citep{yu2024extend}} introduces a method to merge a fine-tuned model with a base model which has undergone additional training. This is in some sense comparable to Param $\Delta$, however, in the case of WIDEN, the base model with additional training has undergone heavy specialization to a particular set of languages.

In contrast, we investigate merging as a way to mitigate the alignment tax. We are the first to show that merging an instruction-tuned model with the base model consistently reveals models that achieve higher accuracy than both parent models and improve calibration compared to the IT model.

\section{Theoretical Interpretation via Linear Mode Connectivity}
\label{sec:appendix_lmc}

We interpret our results through the lens of loss landscape geometry. Instruction tuning typically drives models into ``sharp'' minima characterized by low entropy and high confidence, effectively overfitting to the instruction distribution. In contrast, pre-trained models reside in broader, higher-entropy basins. Because fine-tuned models remain \textit{linearly mode connected} to their initialization~\citep{DBLP:conf/icml/FrankleD0C20}, interpolation effectively moves the weights out of the sharp IT minimum toward the flatter PT basin, a geometric shift known to improve calibration and generalization~\citep{DBLP:conf/uai/IzmailovPGVW18,foret2021sharpnessaware}. Theoretically, this can be viewed as a deterministic approximation of Bayesian Model Averaging~\citep{DBLP:conf/nips/MaddoxIGVW19}, where merging regulates the update strength to prevent the posterior collapse (overconfidence) inherent in standard fine-tuning, effectively recovering the model's prior on uncertainty.

\begin{figure}[ht]
    \centering
    \includegraphics[width=0.95\linewidth]{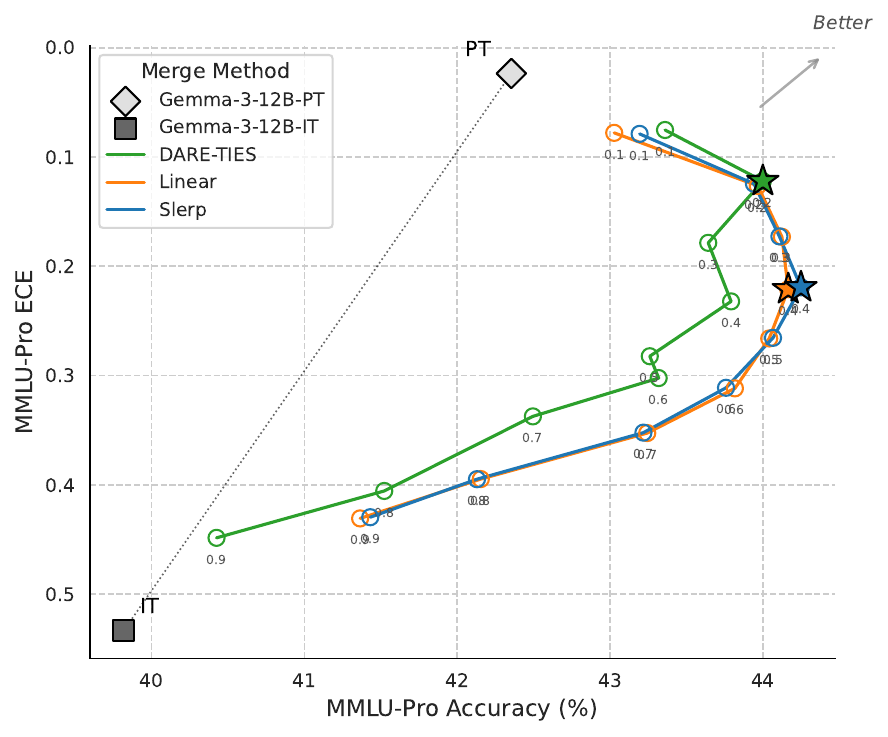}
\caption{Different merging methods (SLERP, Linear, DARE-TIES) trace distinct but similar Pareto-superior paths.}
    \label{fig:robustness_across_merging_method}
\end{figure}

\section{Amplifying Task Vectors Leads to Performance Collapse}
\label{sec:appendix_amplifying}

While merging offers a tunable knob to navigate the alignment-calibration frontier, we also investigated the effect of amplifying a task vector by setting its coefficient $\lambda > 1$. This experiment tests whether one can ``supercharge'' a model's capabilities by pushing its weights further along a specific skill direction. Our findings, detailed in Table~\ref{tab:lambda_degradation_ece}, reveal that this approach is fundamentally destructive. We observe a performance cliff: as $\lambda$ increases beyond 1, model performance enters a sharp and monotonic decline across all evaluation axes. The degradation is particularly catastrophic on benchmarks measuring Helpfulness and complex reasoning. For instance, performance on \texttt{IFEval} plummets from 75.4\% at $\lambda=1.1$ to just 10.3\% at $\lambda=2.0$, and accuracy on \texttt{MATH Lvl 5} collapses from 55.1\% to a near-total failure of 0.6\%. Concurrently, the model's calibration degrades severely; ECE scores on BBH, GPQA, and MMLU-PRO consistently worsen, indicating that the model becomes progressively more miscalibrated and overconfident as its accuracy falls. This demonstrates that amplifying task vectors is not a viable strategy for capability enhancement; instead, it systematically dismantles the model's general capabilities and its grasp on probabilistic uncertainty.

\section{Safety Evaluation of Merged Models}
\label{sec:appendix_safety}

A natural concern with merging instruction-tuned models back toward their base is the potential dilution of safety guardrails. To quantify this, we conduct a sweep across the full interpolation range ($\lambda \in [0.1, 0.9]$) for both Gemma-3-12B and Qwen-2.5-7B using ToxiGen~\citep{hartvigsen-etal-2022-toxigen} (generative toxicity) and WMDP~\citep{DBLP:conf/icml/LiPGYBGLDGMHLJL24} (hazardous knowledge).

Table~\ref{tab:safety_sweep} presents the results. On ToxiGen, safety scores decrease as base weights are introduced, stabilizing around the base model level ($\sim$43--60\%). Crucially, hazardous knowledge scores on WMDP remain relatively flat across the entire sweep (e.g., Qwen stays between 60--62\%). This demonstrates that merging does not trigger a release of latent dangerous capabilities significantly beyond the base model.

More importantly, we argue that in the open-weight ecosystem, the base model represents the ``Safety Floor.'' Since the base model is already publicly available, merging does not introduce a net-new risk to the community. Any merge between a base model and an instruct model is inherently at least as safe as the base model alone.

\begin{table}[h!]
\centering
\setlength{\tabcolsep}{4pt}
\small
\begin{tabular}{l l c c}
\toprule
\textbf{Family} & \textbf{Model / $\lambda$} & \textbf{ToxiGen ($\uparrow$)} & \textbf{WMDP ($\downarrow$)} \\
\midrule
\textbf{Gemma-3} & \textbf{Base (PT)} & 43.2 & 56.2 \\
 & $\lambda = 0.1$ & 43.2 & 57.2 \\
 & $\lambda = 0.2$ & 43.2 & 58.0 \\
 & $\lambda = 0.3$ & 43.2 & 57.8 \\
 & $\lambda = 0.4$ & 43.3 & 58.6 \\
 & $\lambda = 0.5$ & 43.3 & 58.8 \\
 & $\lambda = 0.6$ & 45.9 & 58.7 \\
 & $\lambda = 0.7$ & 54.6 & 58.5 \\
 & $\lambda = 0.8$ & 58.1 & 58.8 \\
 & $\lambda = 0.9$ & 58.2 & 58.3 \\
 & \textbf{Instruct (IT)} & 86.4 & 56.4 \\ 
\midrule
\textbf{Qwen-2.5} & \textbf{Base (PT)} & 60.3 & 60.8 \\
 & $\lambda = 0.1$ & 59.3 & 60.8 \\
 & $\lambda = 0.2$ & 58.6 & 60.7 \\
 & $\lambda = 0.3$ & 58.0 & 60.8 \\
 & $\lambda = 0.4$ & 57.8 & 61.5 \\
 & $\lambda = 0.5$ & 58.5 & 62.1 \\
 & $\lambda = 0.6$ & 58.2 & 62.3 \\
 & $\lambda = 0.7$ & 58.3 & 62.5 \\
 & $\lambda = 0.8$ & 58.0 & 62.2 \\
 & $\lambda = 0.9$ & 58.1 & 62.1 \\
 & \textbf{Instruct (IT)} & 82.8 & 58.4 \\ 
\bottomrule
\end{tabular}
\caption{Safety evaluation across the merge interpolation range. ToxiGen measures safety (higher is better), while WMDP measures hazardous knowledge accuracy (lower is better). Merging does not release latent dangerous capabilities beyond the base model.}
\label{tab:safety_sweep}
\end{table}

\section{Comparison with Alternative Calibration Methods}
\label{sec:appendix_alternatives}

\paragraph{Temperature Scaling.} While Temperature Scaling~\citep{guo2017calibration} is a standard post-hoc calibration technique, it has fundamental limitations for our setting: (1) it is a monotonic transformation that preserves prediction rankings and thus cannot improve accuracy, only ECE; (2) it requires a labeled validation set per task, whereas our method is zero-shot; and (3) it adjusts scalar confidence scores but cannot improve the quality of open-ended generation, as demonstrated by our NoveltyBench results (Figure~\ref{fig:novelty_bench}).

\paragraph{Self-Consistency.} Self-Consistency~\cite{DBLP:conf/iclr/0002WSLCNCZ23} relies on sampling diverse reasoning paths. However, as we diagnose in Section~4, alignment induces mode collapse (extreme confidence inflation). If a model is structurally overconfident in an incorrect answer, Self-Consistency will repeatedly sample the same wrong prediction. Furthermore, it incurs $O(k)$ inference cost, whereas our merged model restores the underlying distribution at $O(1)$ cost.

\scriptsize
\onecolumn
\begin{longtable}{l c | S[table-format=2.2] S[table-format=1.3] | S[table-format=2.2] S[table-format=1.3] | S[table-format=2.2] S[table-format=1.3] | S[table-format=2.2] S[table-format=2.2]}
\caption{Full Performance and Calibration Comparison of Merged Models}
\label{tab:comprehensive_results} \\

\toprule
\textbf{Model} & {\textbf{Type / $\lambda$}} & 
\multicolumn{2}{p{2.5cm}|}{\centering\textbf{BBH}} & 
\multicolumn{2}{p{2.5cm}|}{\centering\textbf{GPQA}} & 
\multicolumn{2}{p{2.5cm}|}{\centering\textbf{MMLU-PRO}} & 
\multicolumn{2}{p{2.5cm}}{\centering\textbf{Additional Benchmarks (Accuracy \%)}} \\
\cmidrule(lr){3-4} \cmidrule(lr){5-6} \cmidrule(lr){7-8} \cmidrule(lr){9-10}
 & & {Acc (\%)} & {ECE} & {Acc (\%)} & {ECE} & {Acc (\%)} & {ECE} & {IFEval} & {MATH L5} \\
\midrule
\endfirsthead

\multicolumn{10}{c}%
{{\tablename\ \thetable{} -- continued from previous page}} \\
\toprule
\textbf{Model} & {\textbf{Type / $\lambda$}} & 
\multicolumn{2}{p{2.5cm}|}{\centering\textbf{BBH}} & 
\multicolumn{2}{p{2.5cm}|}{\centering\textbf{GPQA}} & 
\multicolumn{2}{p{2.5cm}|}{\centering\textbf{MMLU-PRO}} & 
\multicolumn{2}{p{2.5cm}}{\centering\textbf{Additional Benchmarks (Accuracy \%)}} \\
\cmidrule(lr){3-4} \cmidrule(lr){5-6} \cmidrule(lr){7-8} \cmidrule(lr){9-10}
 & & {Acc (\%)} & {ECE} & {Acc (\%)} & {ECE} & {Acc (\%)} & {ECE} & {IFEval} & {MATH L5} \\
\midrule
\endhead

\bottomrule
\endfoot

\endlastfoot

\multicolumn{10}{l}{\textit{\textbf{Gemma-3 12B}}} \\
\midrule
gemma-3-12b-pt & \textit{Base PT} & 54.31 & 0.022 & 34.65 & 0.046 & 42.35 & 0.024 & 19.41 & 16.31 \\
gemma-3-12b-it & \textit{Base IT} & 63.27 & 0.325 & 33.64 & 0.597 & 39.82 & 0.533 & 77.08 & 55.82 \\
\midrule
\multicolumn{10}{l}{\textit{Gemma-3 12B SLERP Merges}} \\
\midrule
gemma3-12b-slerp & 0.1 & 55.89 & 0.034 & 34.82 & 0.087 & 43.19 & 0.079 & 20.89 & 23.79 \\
gemma3-12b-slerp & 0.2 & 57.82 & 0.059 & 35.74 & 0.123 & 43.94 & 0.125 & 26.62 & 26.59 \\
gemma3-12b-slerp & 0.3 & 60.06 & 0.083 & 35.32 & 0.175 & 44.11 & 0.173 & 35.49 & 32.40 \\
gemma3-12b-slerp & 0.4 & 61.08 & 0.112 & 35.32 & 0.224 & 44.25 & 0.219 & 40.67 & 38.75 \\
gemma3-12b-slerp & 0.5 & 61.74 & 0.146 & 35.23 & 0.272 & 44.07 & 0.265 & 47.50 & 42.98 \\
gemma3-12b-slerp & 0.6 & 62.66 & 0.172 & 34.40 & 0.329 & 43.76 & 0.311 & 48.24 & 46.75 \\
gemma3-12b-slerp & 0.7 & 62.63 & 0.202 & 34.90 & 0.368 & 43.22 & 0.352 & 55.45 & 48.49 \\
gemma3-12b-slerp & 0.8 & 62.91 & 0.229 & 33.64 & 0.423 & 42.13 & 0.395 & 67.10 & 52.87 \\
gemma3-12b-slerp & 0.9 & 62.45 & 0.260 & 33.05 & 0.471 & 41.43 & 0.430 & 76.16 & 55.06 \\
\midrule
\multicolumn{10}{l}{\textit{Gemma-3 12B Linear Merges}} \\
\midrule
gemma3-12b-linear & 0.1 & 55.84 & 0.032 & 35.57 & 0.093 & 43.03 & 0.078 & 22.00 & 23.26 \\
gemma3-12b-linear & 0.2 & 57.75 & 0.064 & 35.82 & 0.121 & 43.97 & 0.126 & 26.43 & 26.21 \\
gemma3-12b-linear & 0.3 & 59.76 & 0.085 & 35.07 & 0.178 & 44.12 & 0.173 & 34.75 & 32.55 \\
gemma3-12b-linear & 0.4 & 60.81 & 0.116 & 35.32 & 0.224 & 44.17 & 0.221 & 41.40 & 39.12 \\
gemma3-12b-linear & 0.5 & 61.66 & 0.147 & 35.15 & 0.274 & 44.04 & 0.266 & 46.40 & 43.28 \\
gemma3-12b-linear & 0.6 & 62.85 & 0.170 & 34.82 & 0.326 & 43.82 & 0.312 & 49.17 & 47.21 \\
gemma3-12b-linear & 0.7 & 62.59 & 0.203 & 34.56 & 0.371 & 43.24 & 0.353 & 55.45 & 48.72 \\
gemma3-12b-linear & 0.8 & 63.20 & 0.228 & 34.06 & 0.419 & 42.15 & 0.395 & 67.65 & 51.89 \\
gemma3-12b-linear & 0.9 & 62.49 & 0.259 & 32.80 & 0.476 & 41.36 & 0.431 & 75.05 & 55.89 \\
\midrule
\multicolumn{10}{l}{\textit{Gemma-3 12B DARE-TIES Merges}} \\
\midrule
gemma3-12b-dare\_ties & 0.1 & 55.89 & 0.035 & 35.32 & 0.084 & 43.36 & 0.075 & 20.52 & 24.47 \\
gemma3-12b-dare\_ties & 0.2 & 57.79 & 0.059 & 34.90 & 0.126 & 44.00 & 0.122 & 25.88 & 26.21 \\
gemma3-12b-dare\_ties & 0.3 & 59.16 & 0.089 & 35.32 & 0.174 & 43.64 & 0.179 & 35.30 & 32.25 \\
gemma3-12b-dare\_ties & 0.4 & 60.39 & 0.118 & 35.65 & 0.229 & 43.79 & 0.232 & 48.98 & 37.84 \\
gemma3-12b-dare\_ties & 0.5 & 62.02 & 0.135 & 35.15 & 0.279 & 43.26 & 0.282 & 45.66 & 41.99 \\
gemma3-12b-dare\_ties & 0.6 & 62.51 & 0.172 & 33.81 & 0.327 & 43.32 & 0.302 & 50.09 & 44.86 \\
gemma3-12b-dare\_ties & 0.7 & 62.47 & 0.203 & 33.81 & 0.371 & 42.50 & 0.337 & 56.19 & 49.17 \\
gemma3-12b-dare\_ties & 0.8 & 61.88 & 0.236 & 34.14 & 0.424 & 41.52 & 0.406 & 65.99 & 51.59 \\
gemma3-12b-dare\_ties & 0.9 & 61.34 & 0.291 & 33.56 & 0.463 & 40.43 & 0.448 & 73.75 & 50.60 \\
\midrule
\multicolumn{10}{l}{\textit{\textbf{Gemma-3 27B Family}}} \\
\midrule
gemma-3-27b-pt & \textit{Base PT} & 61.66 & 0.049 & 34.98 & 0.068 & 49.39 & 0.037 & 20.33 & 24.92 \\
gemma-3-27b-it & \textit{Base IT} & 67.21 & 0.302 & 36.24 & 0.590 & 47.80 & 0.478 & 80.59 & 63.14 \\
\midrule
\multicolumn{10}{l}{\textit{Gemma-3 27B SLERP Merges}} \\
\midrule
gemma3-27b-slerp & 0.1 & 64.03 & 0.015 & 36.91 & 0.088 & 50.41 & 0.077 & 22.92 & 33.08 \\
gemma3-27b-slerp & 0.2 & 65.28 & 0.037 & 36.16 & 0.140 & 50.89 & 0.119 & 30.68 & 36.71 \\
gemma3-27b-slerp & 0.3 & 65.86 & 0.075 & 35.49 & 0.192 & 51.44 & 0.156 & 40.67 & 43.81 \\
gemma3-27b-slerp & 0.4 & 66.69 & 0.104 & 37.33 & 0.219 & 51.46 & 0.192 & 48.24 & 48.87 \\
gemma3-27b-slerp & 0.5 & 67.12 & 0.133 & 37.42 & 0.266 & 51.91 & 0.219 & 51.94 & 52.64 \\
gemma3-27b-slerp & 0.6 & 67.23 & 0.161 & 37.84 & 0.305 & 51.11 & 0.254 & 52.31 & 56.50 \\
gemma3-27b-slerp & 0.7 & 67.07 & 0.189 & 38.17 & 0.350 & 50.96 & 0.284 & 63.96 & 58.61 \\
gemma3-27b-slerp & 0.8 & 67.56 & 0.209 & 37.84 & 0.403 & 50.58 & 0.312 & 75.60 & 62.16 \\
gemma3-27b-slerp & 0.9 & 67.52 & 0.232 & 37.08 & 0.455 & 49.78 & 0.341 & 77.63 & 63.14 \\
\midrule
\multicolumn{10}{l}{\textit{Gemma-3 27B Linear Merges}} \\
\midrule
gemma3-27b-linear & 0.1 & 63.79 & 0.017 & 36.74 & 0.088 & 50.23 & 0.078 & 24.58 & 33.16 \\
gemma3-27b-linear & 0.2 & 65.18 & 0.038 & 36.49 & 0.135 & 50.84 & 0.122 & 30.50 & 37.39 \\
gemma3-27b-linear & 0.3 & 66.06 & 0.073 & 35.57 & 0.190 & 51.44 & 0.156 & 39.19 & 43.96 \\
gemma3-27b-linear & 0.4 & 66.55 & 0.106 & 36.74 & 0.224 & 51.55 & 0.192 & 49.91 & 48.34 \\
gemma3-27b-linear & 0.5 & 67.19 & 0.133 & 36.91 & 0.269 & 51.59 & 0.222 & 51.76 & 52.49 \\
gemma3-27b-linear & 0.6 & 67.38 & 0.161 & 38.00 & 0.305 & 51.16 & 0.255 & 53.97 & 55.59 \\
gemma3-27b-linear & 0.7 & 67.28 & 0.188 & 38.17 & 0.349 & 50.82 & 0.285 & 63.03 & 59.14 \\
gemma3-27b-linear & 0.8 & 67.70 & 0.207 & 37.33 & 0.407 & 50.54 & 0.314 & 77.63 & 62.08 \\
gemma3-27b-linear & 0.9 & 67.35 & 0.234 & 36.91 & 0.455 & 49.69 & 0.342 & 78.37 & 61.93 \\
\midrule
\multicolumn{10}{l}{\textit{Gemma-3 27B DARE-TIES Merges}} \\
\midrule
gemma3-27b-dare\_ties & 0.1 & 63.95 & 0.015 & 36.41 & 0.094 & 50.28 & 0.078 & 23.48 & 32.63 \\
gemma3-27b-dare\_ties & 0.2 & 65.32 & 0.035 & 36.16 & 0.141 & 50.83 & 0.121 & 31.42 & 37.76 \\
gemma3-27b-dare\_ties & 0.3 & 66.17 & 0.071 & 35.91 & 0.183 & 51.41 & 0.154 & 39.74 & 43.28 \\
gemma3-27b-dare\_ties & 0.4 & 66.38 & 0.108 & 36.83 & 0.227 & 51.48 & 0.193 & 51.57 & 48.94 \\
gemma3-27b-dare\_ties & 0.5 & 67.25 & 0.130 & 37.25 & 0.267 & 51.21 & 0.227 & 49.35 & 53.02 \\
gemma3-27b-dare\_ties & 0.6 & 67.58 & 0.158 & 37.84 & 0.302 & 50.96 & 0.260 & 58.04 & 55.74 \\
gemma3-27b-dare\_ties & 0.7 & 67.16 & 0.192 & 36.74 & 0.368 & 50.39 & 0.286 & 65.99 & 58.76 \\
gemma3-27b-dare\_ties & 0.8 & 67.44 & 0.211 & 36.07 & 0.424 & 49.73 & 0.322 & 76.89 & 60.50 \\
gemma3-27b-dare\_ties & 0.9 & 66.72 & 0.238 & 38.17 & 0.434 & 48.55 & 0.347 & 78.19 & 63.14 \\
\midrule
\multicolumn{10}{l}{\textit{\textbf{Gemma-3 4B Family}}} \\
\midrule
gemma-3-4b-pt & \textit{PT} & 40.58 & 0.048 & 29.53 & 0.077 & 27.94 & 0.024 & 20.89 & 7.33 \\
gemma-3-4b-it & \textit{IT} & 49.96 & 0.476 & 29.03 & 0.637 & 29.80 & 0.642 & 70.43 & 38.07 \\
\midrule
\multicolumn{10}{l}{\textit{Gemma-3 4B SLERP Merges}} \\
\midrule
gemma3-4b-slerp & 0.1 & 42.20 & 0.038 & 30.12 & 0.109 & 29.23 & 0.063 & 21.81 & 8.91 \\
gemma3-4b-slerp & 0.2 & 43.83 & 0.045 & 29.11 & 0.166 & 29.99 & 0.109 & 22.74 & 11.25 \\
gemma3-4b-slerp & 0.3 & 45.60 & 0.082 & 28.36 & 0.224 & 30.39 & 0.163 & 26.43 & 13.29 \\
gemma3-4b-slerp & 0.4 & 46.78 & 0.128 & 29.03 & 0.275 & 30.56 & 0.215 & 29.57 & 17.67 \\
gemma3-4b-slerp & 0.5 & 47.44 & 0.186 & 27.85 & 0.345 & 30.55 & 0.272 & 35.49 & 21.90 \\
gemma3-4b-slerp & 0.6 & 47.89 & 0.243 & 28.44 & 0.393 & 30.19 & 0.334 & 36.97 & 27.57 \\
gemma3-4b-slerp & 0.7 & 48.08 & 0.298 & 28.27 & 0.446 & 29.70 & 0.399 & 40.48 & 31.19 \\
gemma3-4b-slerp & 0.8 & 48.78 & 0.339 & 28.61 & 0.496 & 28.91 & 0.468 & 49.72 & 37.16 \\
gemma3-4b-slerp & 0.9 & 48.90 & 0.380 & 28.36 & 0.550 & 28.52 & 0.527 & 60.81 & 37.16 \\
\midrule
\multicolumn{10}{l}{\textit{Gemma-3 4B Linear Merges}} \\
\midrule
gemma3-4b-linear & 0.1 & 42.15 & 0.040 & 28.94 & 0.119 & 28.95 & 0.065 & 22.37 & 9.06 \\
gemma3-4b-linear & 0.2 & 44.11 & 0.043 & 28.78 & 0.166 & 30.05 & 0.109 & 22.55 & 11.63 \\
gemma3-4b-linear & 0.3 & 45.50 & 0.083 & 28.27 & 0.226 & 30.44 & 0.161 & 27.36 & 13.60 \\
gemma3-4b-linear & 0.4 & 46.26 & 0.133 & 28.27 & 0.282 & 30.67 & 0.212 & 30.31 & 17.98 \\
gemma3-4b-linear & 0.5 & 47.34 & 0.186 & 28.02 & 0.346 & 30.41 & 0.274 & 34.01 & 21.75 \\
gemma3-4b-linear & 0.6 & 48.00 & 0.242 & 27.60 & 0.401 & 30.15 & 0.333 & 38.08 & 27.27 \\
gemma3-4b-linear & 0.7 & 47.77 & 0.299 & 28.52 & 0.449 & 29.72 & 0.401 & 42.14 & 31.19 \\
gemma3-4b-linear & 0.8 & 48.53 & 0.341 & 27.94 & 0.504 & 28.97 & 0.465 & 49.54 & 36.33 \\
gemma3-4b-linear & 0.9 & 48.76 & 0.381 & 28.69 & 0.546 & 28.38 & 0.527 & 63.77 & 38.67 \\
\midrule
\multicolumn{10}{l}{\textit{Gemma-3 4B DARE-TIES Merges}} \\
\midrule
gemma3-4b-dare\_ties & 0.1 & 42.25 & 0.042 & 29.78 & 0.110 & 29.28 & 0.064 & 21.81 & 8.76 \\
gemma3-4b-dare\_ties & 0.2 & 43.57 & 0.043 & 30.29 & 0.142 & 29.85 & 0.103 & 24.40 & 10.95 \\
gemma3-4b-dare\_ties & 0.3 & 44.92 & 0.085 & 28.86 & 0.227 & 30.11 & 0.169 & 28.10 & 13.97 \\
gemma3-4b-dare\_ties & 0.4 & 45.93 & 0.134 & 27.94 & 0.286 & 30.65 & 0.197 & 32.53 & 16.47 \\
gemma3-4b-dare\_ties & 0.5 & 46.52 & 0.202 & 27.94 & 0.345 & 29.72 & 0.293 & 33.09 & 21.07 \\
gemma3-4b-dare\_ties & 0.6 & 46.83 & 0.256 & 27.52 & 0.430 & 29.60 & 0.360 & 35.86 & 24.55 \\
gemma3-4b-dare\_ties & 0.7 & 47.46 & 0.303 & 26.85 & 0.461 & 28.61 & 0.412 & 41.59 & 29.08 \\
gemma3-4b-dare\_ties & 0.8 & 48.88 & 0.339 & 28.10 & 0.511 & 28.77 & 0.412 & 46.58 & 32.93 \\
gemma3-4b-dare\_ties & 0.9 & 47.65 & 0.370 & 26.26 & 0.561 & 26.97 & 0.503 & 61.92 & 33.16 \\
\midrule
\multicolumn{10}{l}{\textit{\textbf{Qwen 2.5 1.5B}}} \\
\midrule
Qwen2.5-1.5B & \textit{PT} & 40.50 & 0.105 & 28.27 & 0.114 & 28.73 & 0.055 & 22.74 & 8.91 \\
Qwen2.5-1.5B-Instruct & \textit{IT} & 42.37 & 0.248 & 26.17 & 0.244 & 28.08 & 0.326 & 41.22 & 21.75 \\
\midrule
\multicolumn{10}{l}{\textit{Qwen 2.5 1.5B SLERP Merges}} \\
\midrule
qwen2.5-1.5b-slerp & 0.1 & 40.51 & 0.112 & 29.70 & 0.107 & 28.38 & 0.065 & 22.00 & 8.99 \\
qwen2.5-1.5b-slerp & 0.2 & 41.24 & 0.115 & 29.61 & 0.122 & 28.62 & 0.076 & 22.74 & 10.05 \\
qwen2.5-1.5b-slerp & 0.3 & 41.83 & 0.123 & 28.52 & 0.150 & 28.75 & 0.089 & 22.37 & 9.89 \\
qwen2.5-1.5b-slerp & 0.4 & 42.01 & 0.130 & 28.61 & 0.162 & 28.81 & 0.097 & 24.77 & 10.73 \\
qwen2.5-1.5b-slerp & 0.5 & 41.94 & 0.145 & 29.11 & 0.183 & 28.98 & 0.113 & 24.77 & 10.57 \\
qwen2.5-1.5b-slerp & 0.6 & 42.21 & 0.154 & 29.03 & 0.191 & 29.24 & 0.125 & 26.25 & 10.35 \\
qwen2.5-1.5b-slerp & 0.7 & 42.30 & 0.162 & 28.02 & 0.211 & 29.31 & 0.136 & 25.88 & 10.20 \\
qwen2.5-1.5b-slerp & 0.8 & 42.54 & 0.173 & 28.44 & 0.222 & 29.35 & 0.152 & 26.80 & 10.57 \\
qwen2.5-1.5b-slerp & 0.9 & 42.13 & 0.191 & 28.61 & 0.235 & 29.13 & 0.172 & 26.62 & 10.27 \\
\midrule
\multicolumn{10}{l}{\textit{Qwen 2.5 1.5B Linear Merges}} \\
\midrule
qwen2.5-1.5b-linear & 0.1 & 40.83 & 0.112 & 29.45 & 0.113 & 28.72 & 0.064 & 22.37 & 10.05 \\
qwen2.5-1.5b-linear & 0.2 & 41.52 & 0.115 & 29.78 & 0.123 & 28.73 & 0.075 & 19.04 & 9.29 \\
qwen2.5-1.5b-linear & 0.3 & 41.61 & 0.124 & 28.86 & 0.146 & 28.64 & 0.089 & 21.63 & 9.82 \\
qwen2.5-1.5b-linear & 0.4 & 42.32 & 0.128 & 29.11 & 0.159 & 28.81 & 0.101 & 24.40 & 9.37 \\
qwen2.5-1.5b-linear & 0.5 & 41.94 & 0.145 & 29.11 & 0.183 & 28.98 & 0.113 & 19.96 & 10.73 \\
qwen2.5-1.5b-linear & 0.6 & 42.15 & 0.154 & 28.36 & 0.196 & 29.23 & 0.123 & 26.06 & 10.12 \\
qwen2.5-1.5b-linear & 0.7 & 42.09 & 0.166 & 28.69 & 0.205 & 29.30 & 0.134 & 27.36 & 10.88 \\
qwen2.5-1.5b-linear & 0.8 & 42.46 & 0.173 & 28.94 & 0.218 & 29.32 & 0.150 & 23.11 & 10.27 \\
qwen2.5-1.5b-linear & 0.9 & 42.27 & 0.188 & 28.10 & 0.236 & 29.26 & 0.163 & 24.95 & 10.50 \\
\midrule
\multicolumn{10}{l}{\textit{Qwen 2.5 1.5B DARE-TIES Merges}} \\
\midrule
qwen2.5-1.5b-dare\_ties & 0.1 & 40.86 & 0.108 & 29.53 & 0.110 & 28.56 & 0.064 & 22.55 & 10.05 \\
qwen2.5-1.5b-dare\_ties & 0.2 & 41.05 & 0.117 & 29.53 & 0.122 & 28.57 & 0.076 & 23.48 & 9.29 \\
qwen2.5-1.5b-dare\_ties & 0.3 & 41.75 & 0.123 & 29.36 & 0.146 & 28.73 & 0.087 & 22.74 & 10.57 \\
qwen2.5-1.5b-dare\_ties & 0.4 & 42.02 & 0.129 & 29.19 & 0.159 & 28.80 & 0.096 & 24.21 & 10.57 \\
qwen2.5-1.5b-dare\_ties & 0.5 & 42.21 & 0.144 & 28.86 & 0.181 & 29.01 & 0.115 & 26.06 & 10.50 \\
qwen2.5-1.5b-dare\_ties & 0.6 & 42.13 & 0.155 & 28.44 & 0.193 & 29.25 & 0.122 & 26.80 & 10.65 \\
qwen2.5-1.5b-dare\_ties & 0.7 & 42.11 & 0.166 & 28.36 & 0.211 & 29.45 & 0.137 & 24.21 & 9.97 \\
qwen2.5-1.5b-dare\_ties & 0.8 & 42.11 & 0.178 & 28.86 & 0.221 & 29.29 & 0.155 & 27.36 & 9.37 \\
qwen2.5-1.5b-dare\_ties & 0.9 & 42.08 & 0.187 & 28.52 & 0.233 & 29.44 & 0.164 & 26.62 & 9.29 \\
\midrule
\multicolumn{10}{l}{\textit{\textbf{Qwen 2.5 3B}}} \\
\midrule
Qwen2.5-3B & \textit{PT} & 46.38 & 0.102 & 28.36 & 0.183 & 32.12 & 0.044 & 20.89 & 15.94 \\
Qwen2.5-3B-Instruct & \textit{IT} & 46.59 & 0.455 & 28.78 & 0.347 & 32.77 & 0.469 & 58.04 & 37.54 \\
\midrule
\multicolumn{10}{l}{\textit{Qwen 2.5 3B SLERP Merges}} \\
\midrule
qwen2.5-3b-slerp & 0.1 & 46.50 & 0.122 & 28.61 & 0.188 & 32.71 & 0.052 & 23.11 & 15.26 \\
qwen2.5-3b-slerp & 0.2 & 47.37 & 0.148 & 28.36 & 0.213 & 33.96 & 0.061 & 28.10 & 16.62 \\
qwen2.5-3b-slerp & 0.3 & 47.77 & 0.177 & 29.03 & 0.232 & 34.94 & 0.079 & 36.04 & 18.20 \\
qwen2.5-3b-slerp & 0.4 & 47.63 & 0.206 & 27.85 & 0.265 & 35.58 & 0.094 & 36.97 & 18.20 \\
qwen2.5-3b-slerp & 0.5 & 47.80 & 0.240 & 29.36 & 0.275 & 36.35 & 0.122 & 39.37 & 20.47 \\
qwen2.5-3b-slerp & 0.6 & 47.98 & 0.277 & 29.19 & 0.302 & 36.94 & 0.155 & 42.51 & 19.11 \\
qwen2.5-3b-slerp & 0.7 & 47.96 & 0.307 & 29.61 & 0.319 & 37.36 & 0.182 & 48.43 & 20.39 \\
qwen2.5-3b-slerp & 0.8 & 47.77 & 0.348 & 30.03 & 0.344 & 37.67 & 0.224 & 45.29 & 21.15 \\
qwen2.5-3b-slerp & 0.9 & 47.94 & 0.382 & 29.45 & 0.379 & 37.82 & 0.267 & 48.24 & 25.30 \\
\midrule
\multicolumn{10}{l}{\textit{Qwen 2.5 3B Linear Merges}} \\
\midrule
qwen2.5-3b-linear & 0.1 & 46.83 & 0.125 & 28.19 & 0.196 & 33.04 & 0.054 & 19.59 & 15.63 \\
qwen2.5-3b-linear & 0.2 & 47.02 & 0.155 & 28.27 & 0.217 & 33.80 & 0.066 & 28.84 & 16.47 \\
qwen2.5-3b-linear & 0.3 & 47.56 & 0.180 & 27.60 & 0.247 & 35.06 & 0.079 & 34.94 & 17.82 \\
qwen2.5-3b-linear & 0.4 & 47.68 & 0.211 & 28.27 & 0.265 & 35.67 & 0.098 & 38.82 & 18.20 \\
qwen2.5-3b-linear & 0.5 & 47.80 & 0.240 & 29.36 & 0.275 & 36.35 & 0.122 & 39.37 & 20.47 \\
qwen2.5-3b-linear & 0.6 & 47.79 & 0.274 & 28.86 & 0.303 & 36.76 & 0.149 & 29.02 & 18.43 \\
qwen2.5-3b-linear & 0.7 & 47.91 & 0.309 & 29.11 & 0.326 & 37.18 & 0.186 & 30.31 & 17.22 \\
qwen2.5-3b-linear & 0.8 & 47.53 & 0.348 & 29.61 & 0.347 & 37.52 & 0.218 & 46.95 & 21.45 \\
qwen2.5-3b-linear & 0.9 & 47.84 & 0.375 & 29.70 & 0.372 & 37.67 & 0.260 & 50.09 & 21.15 \\
\midrule
\multicolumn{10}{l}{\textit{Qwen 2.5 3B DARE-TIES Merges}} \\
\midrule
qwen2.5-3b-dare\_ties & 0.1 & 46.57 & 0.122 & 28.36 & 0.190 & 32.78 & 0.051 & 24.77 & 17.07 \\
qwen2.5-3b-dare\_ties & 0.2 & 47.16 & 0.147 & 27.85 & 0.217 & 33.68 & 0.064 & 28.47 & 17.60 \\
qwen2.5-3b-dare\_ties & 0.3 & 47.58 & 0.178 & 28.36 & 0.237 & 34.83 & 0.079 & 35.12 & 17.82 \\
qwen2.5-3b-dare\_ties & 0.4 & 47.56 & 0.202 & 28.19 & 0.258 & 35.38 & 0.094 & 36.97 & 18.35 \\
qwen2.5-3b-dare\_ties & 0.5 & 47.86 & 0.246 & 29.36 & 0.279 & 36.49 & 0.127 & 41.77 & 19.11 \\
qwen2.5-3b-dare\_ties & 0.6 & 47.93 & 0.274 & 29.45 & 0.297 & 37.02 & 0.148 & 42.70 & 19.41 \\
qwen2.5-3b-dare\_ties & 0.7 & 47.72 & 0.316 & 29.28 & 0.329 & 37.21 & 0.192 & 47.69 & 18.43 \\
qwen2.5-3b-dare\_ties & 0.8 & 47.87 & 0.350 & 30.03 & 0.346 & 37.71 & 0.226 & 45.66 & 22.05 \\
qwen2.5-3b-dare\_ties & 0.9 & 47.75 & 0.375 & 29.19 & 0.373 & 37.81 & 0.254 & 50.09 & 21.45 \\
\midrule
\multicolumn{10}{l}{\textit{\textbf{Qwen 2.5 7B}}} \\
\midrule
Qwen2.5-7B & \textit{PT} & 53.67 & 0.097 & 32.30 & 0.133 & 43.55 & 0.063 & 29.39 & 22.58 \\
Qwen2.5-7B-Instruct & \textit{IT} & 53.72 & 0.384 & 30.29 & 0.484 & 43.07 & 0.451 & 71.35 & 49.02 \\
\midrule
\multicolumn{10}{l}{\textit{Qwen 2.5 7B SLERP Merges}} \\
\midrule
qwen2.5-7b-slerp & 0.1 & 53.95 & 0.119 & 31.38 & 0.167 & 43.97 & 0.074 & 31.61 & 23.64 \\
qwen2.5-7b-slerp & 0.2 & 54.50 & 0.147 & 32.63 & 0.174 & 44.58 & 0.090 & 35.49 & 25.30 \\
qwen2.5-7b-slerp & 0.3 & 54.71 & 0.179 & 32.05 & 0.201 & 44.88 & 0.110 & 40.67 & 26.28 \\
qwen2.5-7b-slerp & 0.4 & 54.83 & 0.205 & 31.96 & 0.225 & 45.10 & 0.126 & 44.92 & 26.96 \\
qwen2.5-7b-slerp & 0.5 & 55.27 & 0.231 & 31.96 & 0.252 & 45.48 & 0.148 & 51.39 & 28.70 \\
qwen2.5-7b-slerp & 0.6 & 55.13 & 0.262 & 32.05 & 0.277 & 45.53 & 0.171 & 53.42 & 29.98 \\
qwen2.5-7b-slerp & 0.7 & 55.18 & 0.285 & 32.47 & 0.298 & 45.82 & 0.192 & 55.82 & 32.25 \\
qwen2.5-7b-slerp & 0.8 & 55.39 & 0.308 & 32.55 & 0.324 & 45.50 & 0.231 & 58.04 & 35.50 \\
qwen2.5-7b-slerp & 0.9 & 55.51 & 0.329 & 32.47 & 0.356 & 44.98 & 0.277 & 56.75 & 35.80 \\
\midrule
\multicolumn{10}{l}{\textit{Qwen 2.5 7B Linear Merges}} \\
\midrule
qwen2.5-7b-linear & 0.1 & 54.02 & 0.123 & 32.38 & 0.156 & 44.00 & 0.076 & 31.05 & 23.94 \\
qwen2.5-7b-linear & 0.2 & 54.43 & 0.150 & 31.71 & 0.178 & 44.77 & 0.089 & 23.11 & 25.60 \\
qwen2.5-7b-linear & 0.3 & 55.04 & 0.175 & 32.55 & 0.194 & 45.01 & 0.107 & 43.25 & 25.91 \\
qwen2.5-7b-linear & 0.4 & 55.01 & 0.204 & 31.54 & 0.231 & 45.20 & 0.127 & 26.99 & 28.40 \\
qwen2.5-7b-linear & 0.5 & 55.16 & 0.232 & 32.13 & 0.248 & 45.53 & 0.147 & 50.09 & 29.00 \\
qwen2.5-7b-linear & 0.6 & 55.32 & 0.257 & 32.30 & 0.276 & 45.63 & 0.169 & 32.72 & 30.51 \\
qwen2.5-7b-linear & 0.7 & 55.39 & 0.282 & 32.13 & 0.300 & 45.76 & 0.192 & 34.01 & 31.95 \\
qwen2.5-7b-linear & 0.8 & 55.13 & 0.309 & 31.96 & 0.333 & 45.67 & 0.225 & 36.41 & 33.46 \\
qwen2.5-7b-linear & 0.9 & 55.13 & 0.330 & 32.21 & 0.352 & 45.19 & 0.266 & 56.93 & 34.59 \\
\midrule
\multicolumn{10}{l}{\textit{Qwen 2.5 7B DARE-TIES Merges}} \\
\midrule
qwen2.5-7b-dare\_ties & 0.2 & 54.42 & 0.147 & 31.54 & 0.177 & 44.56 & 0.089 & 35.30 & 25.15 \\
qwen2.5-7b-dare\_ties & 0.3 & 54.87 & 0.177 & 32.80 & 0.191 & 45.05 & 0.106 & 41.77 & 26.96 \\
qwen2.5-7b-dare\_ties & 0.4 & 54.97 & 0.202 & 31.80 & 0.229 & 45.02 & 0.125 & 42.88 & 27.95 \\
qwen2.5-7b-dare\_ties & 0.5 & 55.08 & 0.236 & 31.96 & 0.256 & 45.47 & 0.149 & 51.39 & 29.08 \\
qwen2.5-7b-dare\_ties & 0.6 & 55.25 & 0.260 & 32.30 & 0.272 & 45.56 & 0.171 & 51.39 & 29.38 \\
qwen2.5-7b-dare\_ties & 0.7 & 55.41 & 0.284 & 32.05 & 0.303 & 45.57 & 0.199 & 56.38 & 31.87 \\
qwen2.5-7b-dare\_ties & 0.8 & 55.22 & 0.311 & 32.13 & 0.330 & 45.57 & 0.233 & 56.38 & 33.61 \\
qwen2.5-7b-dare\_ties & 0.9 & 55.23 & 0.329 & 32.30 & 0.355 & 45.20 & 0.265 & 56.38 & 34.89 \\
\midrule
\multicolumn{10}{l}{\textit{\textbf{Llama-3.1-8B}}} \\
\midrule
Meta-Llama-3.1-8B & \textit{Base PT} & 46.48 & 0.025 & 31.46 & 0.058 & 32.72 & 0.059 & 7.76 & 5.89 \\
Meta-Llama-3.1-8B-Instruct & \textit{Base IT} & 50.69 & 0.198 & 29.11 & 0.320 & 37.72 & 0.191 & 73.01 & 19.86 \\
\midrule
\multicolumn{10}{l}{\textit{Llama-3.1-8B SLERP Merges}} \\
\midrule
llama-3.1-8b-slerp & 0.1 & 47.77 & 0.025 & 31.63 & 0.059 & 33.56 & 0.055 & 14.42 & 6.27 \\
llama-3.1-8b-slerp & 0.2 & 48.53 & 0.024 & 31.12 & 0.065 & 34.18 & 0.056 & 14.79 & 7.63 \\
llama-3.1-8b-slerp & 0.3 & 48.85 & 0.025 & 30.96 & 0.071 & 35.00 & 0.052 & 18.85 & 8.99 \\
llama-3.1-8b-slerp & 0.4 & 49.64 & 0.028 & 31.54 & 0.068 & 36.05 & 0.048 & 21.07 & 9.74 \\
llama-3.1-8b-slerp & 0.5 & 50.11 & 0.036 & 31.12 & 0.075 & 36.48 & 0.049 & 23.66 & 11.48 \\
llama-3.1-8b-slerp & 0.6 & 50.41 & 0.045 & 31.80 & 0.072 & 36.94 & 0.055 & 25.69 & 11.63 \\
llama-3.1-8b-slerp & 0.7 & 50.48 & 0.058 & 32.97 & 0.065 & 37.48 & 0.054 & 28.84 & 12.99 \\
llama-3.1-8b-slerp & 0.8 & 50.70 & 0.067 & 32.80 & 0.071 & 37.57 & 0.059 & 32.53 & 15.26 \\
llama-3.1-8b-slerp & 0.9 & 51.02 & 0.074 & 32.38 & 0.080 & 37.72 & 0.064 & 39.19 & 15.26 \\
\midrule
\multicolumn{10}{l}{\textit{Llama-3.1-8B DARE-TIES Merges}} \\
\midrule
llama-3.1-8b-dare\_ties & 0.1 & 47.89 & 0.021 & 30.79 & 0.066 & 33.62 & 0.055 & 28.81 & 6.34 \\
llama-3.1-8b-dare\_ties & 0.2 & 48.29 & 0.024 & 31.71 & 0.059 & 34.31 & 0.054 & 29.38 & 6.72 \\
llama-3.1-8b-dare\_ties & 0.3 & 48.88 & 0.027 & 31.04 & 0.070 & 34.98 & 0.052 & 30.37 & 9.37 \\
llama-3.1-8b-dare\_ties & 0.4 & 49.49 & 0.031 & 31.21 & 0.072 & 35.91 & 0.049 & 31.06 & 10.12 \\
llama-3.1-8b-dare\_ties & 0.5 & 50.18 & 0.034 & 31.12 & 0.074 & 36.32 & 0.054 & 31.90 & 11.03 \\
llama-3.1-8b-dare\_ties & 0.6 & 49.92 & 0.051 & 32.38 & 0.066 & 36.98 & 0.054 & 32.85 & 11.86 \\
llama-3.1-8b-dare\_ties & 0.7 & 49.96 & 0.060 & 32.47 & 0.068 & 37.43 & 0.053 & 33.85 & 12.92 \\
llama-3.1-8b-dare\_ties & 0.8 & 50.54 & 0.066 & 32.72 & 0.070 & 37.87 & 0.057 & 35.12 & 14.20 \\
llama-3.1-8b-dare\_ties & 0.9 & 51.14 & 0.070 & 32.05 & 0.086 & 37.42 & 0.070 & 35.64 & 15.03 \\

\bottomrule
\end{longtable}

\twocolumn

\begin{table*}[ht]
\centering
\resizebox{\linewidth}{!}{%
\begin{tabular}{l c | S[table-format=2.2] S[table-format=1.3] | S[table-format=2.2] S[table-format=1.3] | S[table-format=2.2] S[table-format=1.3] | S[table-format=2.2] S[table-format=2.2]}
\toprule
\textbf{Model} & {\textbf{$\lambda$}} & \multicolumn{2}{c|}{\textbf{BBH}} & \multicolumn{2}{c|}{\textbf{GPQA}} & \multicolumn{2}{c|}{\textbf{MMLU-PRO}} & \multicolumn{2}{c}{\textbf{Other Benchmarks}} \\
\cmidrule(lr){3-4} \cmidrule(lr){5-6} \cmidrule(lr){7-8} \cmidrule(lr){9-10}
 & & {Acc (\%)} & {ECE} & {Acc (\%)} & {ECE} & {Acc (\%)} & {ECE} & {IFEval (\%)} & {MATH L5 (\%)} \\
\midrule
\multicolumn{10}{l}{\textit{Reference Models}} \\
\midrule
gemma-3-12b-pt & \textit{Base PT} & 54.31 & 0.022 & 34.65 & 0.046 & 42.35 & 0.024 & 19.41 & 16.31 \\
gemma-3-12b-it & \textit{Base IT} & 63.27 & 0.325 & 33.64 & 0.597 & 39.82 & 0.533 & 77.08 & 55.82 \\
\midrule
\multicolumn{10}{l}{\textit{Arithmetic Task Vector Applied to PT Model}} \\
\midrule
task arithmetic & 1.1 & 62.77 & 0.339 & 32.55 & 0.622 & 38.36 & 0.548 & 75.42 & 55.06 \\
task arithmetic & 1.2 & 61.73 & 0.354 & 32.30 & 0.635 & 36.69 & 0.565 & 71.35 & 52.79 \\
task arithmetic & 1.4 & 59.02 & 0.386 & 30.29 & 0.661 & 32.45 & 0.591 & 64.51 & 42.22 \\
task arithmetic & 1.5 & 57.15 & 0.401 & 29.78 & 0.664 & 29.46 & 0.608 & 58.78 & 33.91 \\
task arithmetic & 1.6 & 53.72 & 0.433 & 29.19 & 0.674 & 26.02 & 0.624 & 50.46 & 21.37 \\
task arithmetic & 1.7 & 48.98 & 0.478 & 28.44 & 0.683 & 22.17 & 0.655 & 43.62 & 10.35 \\
task arithmetic & 1.8 & 43.62 & 0.529 & 27.85 & 0.678 & 17.84 & 0.670 & 32.53 & 2.42 \\
task arithmetic & 1.9 & 38.26 & 0.573 & 23.99 & 0.697 & 13.83 & 0.659 & 19.78 & 1.44 \\
task arithmetic & 2.0 & 31.66 & 0.612 & 24.66 & 0.669 & 11.64 & 0.672 & 10.35 & 0.60 \\
\bottomrule
\end{tabular}
}
\footnotesize{\\ \textit{Note:} PT refers to the Pre-trained base model; IT refers to the post-instruction-tuned model. ECE (Expected Calibration Error) is a measure of miscalibration where higher values are worse.}

\caption{Performance and Calibration Degradation when amplifying an arithmetic task vector (\(\lambda > 1\)) applied to the Gemma-3-12B-PT model. Extrapolating beyond the instruction-tuned model leads to a catastrophic and monotonic decline in performance and calibration across all benchmarks.}
\label{tab:lambda_degradation_ece}

\end{table*}

\end{document}